\documentclass{article}

\PassOptionsToPackage{numbers, compress}{natbib}
\usepackage[main, final]{neurips_2026}

\usepackage[utf8]{inputenc} % allow utf-8 input
\usepackage[T1]{fontenc}    % use 8-bit T1 fonts
\usepackage[pagebackref=true,colorlinks,linkcolor=blue,citecolor=brown]{hyperref}
\usepackage{url}            % simple URL typesetting
\usepackage{booktabs}       % professional-quality tables
\usepackage{amsfonts}       % blackboard math symbols
\usepackage{nicefrac}       % compact symbols for 1/2, etc.
\usepackage{microtype}      % microtypography
\usepackage{xcolor}         % colors
\usepackage{graphicx}
\usepackage{subcaption}

\usepackage{tabularx} 
\usepackage{amsmath}
\usepackage{amssymb}
\usepackage{mathtools}
\usepackage{amsthm}
\usepackage{bbm}
\usepackage{etoc}
\etocsettagdepth{mainpaper}{none}
\etocsettagdepth{appendix}{subsection}
\usepackage{makecell}
\theoremstyle{plain}

\theoremstyle{definition}

\theoremstyle{remark}

\title{Sparse Reward Subsystem in Large Language Models}

\author{%
  Guowei Xu \\
  Tsinghua University \\
  \And
  Mert Yuksekgonul \\
  Stanford University \\
  \And
  James Zou \\
  Stanford University \\
}

\begin{document}

\maketitle

\etocdepthtag.toc{mainpaper}

\begin{abstract}
Recent studies show that LLM hidden states encode reward-related information, such as answer correctness and model confidence. However, existing approaches typically fit black-box probes on the full hidden states, offering little insight into how this information is structured across neurons. In this paper, we show that reward-related information is concentrated in a sparse subset of neurons. Using simple probing, we identify two types of neurons: value neurons, whose activations predict state value, and dopamine neurons, whose activations encode step-level temporal difference (TD) errors. Together, these neurons form a sparse reward subsystem within LLM hidden states. These names are drawn by analogy with neuroscience, where value neurons and dopamine neurons in the biological reward subsystem also encode value and reward prediction errors, respectively. We demonstrate that value neurons are robust and transferable across diverse datasets and models, and provide causal evidence that they encode reward-related information. Finally, we show applications of the reward subsystem: value neurons serve as effective predictors of model confidence, and dopamine neurons can function as a process reward model (PRM) to guide inference-time search. Our code is available at \url{https://github.com/zou-group/reward_subsystem}.
\end{abstract}

\section{Introduction}

In recent years, the reasoning capabilities of large language models (LLMs) have achieved significant breakthroughs, even surpassing humans in various mathematical and scientific tasks~\citep{gpt5,deepseekr1}. Parallel to the continuous pursuit of higher performance, researchers seek to understand the underlying reasoning mechanisms of these models. Recent studies reveal that LLM hidden states encode rich reward-related information, which can be leveraged for weighted learning~\citep{oh2025housecardsmassiveweights}, predicting confidence and answer correctness~\citep{gekhman2025insideout,latentthinking}, detecting hallucinations~\citep{10.24963/ijcai.2025/929}, and performing implicit reasoning~\citep{chen2024stateshiddenhiddenstates}.

However, these approaches typically fit black-box probes on the full hidden states~\citep{belinkov-2022-probing}: they demonstrate that reward-related information can be decoded, but do not characterize \emph{how} this information is structured within the hidden states. Understanding this matters both for mechanistic interpretability, as it clarifies whether reward signals are sparsely encoded, and for practical applications, since identifying a minimal set of responsible neurons can enable more lightweight and targeted applications.

In this paper, we find that the ability of LLM hidden states to predict reward-related information can be attributed to the existence of a small subset of neurons. 
Using simple probing, we identify two types of neurons: \emph{value neurons}, whose activations predict the expected value of the current state, and \emph{dopamine neurons}, whose activations predict the TD error at each reasoning step. Together, these neurons form what we term a \emph{reward subsystem} within LLM hidden states.
These names are drawn by analogy with neuroscience, where biological value neurons represent the subjective value of stimuli during decision-making~\citep{tremblay1999relative, padoa2006neurons} and dopamine neurons encode reward prediction errors~\citep{schultz1998predictive,td_bio,distributionaldopamine, mesolimbic}.

\begin{figure*}[t]
    \centering
    \includegraphics[width=\textwidth]{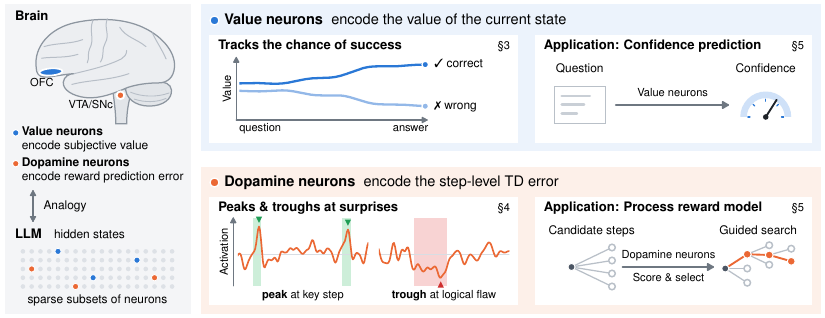}
    \caption{Overview of the sparse reward subsystem in LLMs.
    The left panel illustrates a computational analogy with
    biological value and dopamine neurons.
    Value neurons encode the value of the current state,
    while dopamine neurons encode step-level TD errors.
    The signal illustrations show value estimates tracking
    the chance of success (\S\ref{sec:value}) and dopamine-neuron activations
    exhibiting peaks and troughs at unexpected progress
    and reasoning errors (\S\ref{sec:dopamine}).
    We apply value neurons to confidence prediction
    (\S\ref{sec:app_confidence}) and dopamine neurons to
    process-reward-guided search (\S\ref{sec:app_prm}).}
    \label{fig:reward-subsystem}
\end{figure*}

We first provide motivation for why reward-related information may be sparse in LLM hidden states. Then, for value neurons, we show via pruning experiments that less than 1\% of neurons suffice to predict state value, and establish their specificity as reward-related neurons through intervention experiments. For dopamine neurons, we show via analogous pruning that a similarly sparse subset encodes step-level TD errors, and visualize their activation patterns during reasoning, confirming that they exhibit peaks when the model makes unexpected progress and troughs when the model encounters errors. Finally, we show practical applications of these neurons: value neurons serve as effective predictors of model confidence, and dopamine neurons can function as a process reward model (PRM) to guide inference-time search. Figure~\ref{fig:reward-subsystem} provides an overview
of the reward subsystem and its applications.

In summary, our key contributions are:

\begin{enumerate}
\item We identify sparse value neurons that predict value information, establish their specificity as reward-related neurons through controlled interventions, and demonstrate their robustness across datasets, models, and layers. 
\item We identify sparse dopamine neurons that encode step-level TD errors, with activation patterns consistent with reward prediction error signals. 
\item We highlight applications of these neurons: value neurons can predict model confidence prior to generation, and dopamine neurons can function as an internal process reward model. 
\end{enumerate}

\section{Motivation}
\label{sec:theoretical_motivation}

This section provides motivation for two questions: why autoregressive LLMs may contain neurons that can predict value and TD-error, and why such neurons may appear sparse. The arguments below are intended as motivation, not as rigorous proof. 

\subsection{Why may reward subsystem exist in LLMs?}
\label{sec:theory_existence}

We view an autoregressive LLM $\mathcal{M}$ as a policy over partial reasoning trajectories. The initial state is $s_0$. At generation step $t$, the model samples an action $a_t \sim \mathcal{M}(\cdot \mid s_t)$ and updates the state as $s_{t+1}=s_t\oplus a_t$. Let $r(s_T)\in\{0,1\}$ denote the terminal reward after the model completes a response. 
For a fixed model $\mathcal{M}$, the value of a partial reasoning step can be written as:

\begin{equation}
V^{\mathcal{M}}(s_t)=\mathbb{E}_{\mathcal{M}}\left[ r(s_T) \mid s_t \right],
\label{eq:model_policy_value}
\end{equation}

which is the expected probability that continuing generation from $s_t$ will eventually yield a correct answer.

A useful motivation comes from maximum-entropy reinforcement learning. In the soft-optimal setting, the optimal policy and optimal soft $Q$-function satisfy
\begin{equation}
\pi^*(a \mid s)
=
\frac{\exp(Q^*(s,a)/\tau)}{Z(s)},
\qquad
Z(s)=\sum_{a'} \exp(Q^*(s,a')/\tau),
\label{eq:maxent_policy}
\end{equation}
or equivalently,
\begin{equation}
\log \pi^*(a \mid s)
=
\frac{Q^*(s,a)}{\tau}
-
\log Z(s) = \frac{Q^*(s,a)-V^*(s)}{\tau}.
\label{eq:maxent_log_policy}
\end{equation}
Here, $\tau>0$ is the temperature and
$V^*(s)$ is the optimal \emph{soft} value
function.
These equations illustrate the relation between policy and
value in an idealized soft-optimal setting. Our intuition
is that a sufficiently capable policy may implicitly encode
value-related information. Since LLMs are reasonably capable
on the tasks we study, we investigate whether their internal
representations encode information about the current-policy
value.

\subsection{Why may these neurons be sparse?}
\label{sec:theory_sparsity}

\paragraph{Feasibility.}
Information theory shows that a small number of coordinates can, in principle, suffice to encode value information. Let $R=r(s_T)$ denote the binary terminal reward obtained by completing a trajectory from the current state, and let $h=h(s_t,l)$ denote the hidden state at layer $l$. For predicting $R$ from $h$, the posterior probability
\begin{equation}
\eta(h)=P(R=1\mid h)
\label{eq:posterior_eta}
\end{equation}
is a sufficient scalar summary for binary prediction. By Bayes' rule,
\begin{equation}
P(R=1 \mid h)
=
\frac{p(h\mid R=1)P(R=1)}
{p(h\mid R=1)P(R=1)+p(h\mid R=0)P(R=0)}.
\label{eq:bayes_reward}
\end{equation}
Equivalently,
\begin{equation}
P(R=1 \mid h)
=
\sigma\left(
\log \frac{p(h\mid R=1)}{p(h\mid R=0)}
+
\log \frac{P(R=1)}{P(R=0)}
\right),
\label{eq:bayes_log_ratio}
\end{equation}
where $\sigma$ is the sigmoid function. Thus, for Bayes-optimal prediction of the binary terminal reward, the relevant information in $h$ can in principle be summarized by the scalar log-likelihood ratio
\begin{equation}
T(h)=\log \frac{p(h\mid R=1)}{p(h\mid R=0)}.
\label{eq:log_likelihood_ratio}
\end{equation}
This argument shows that the reward-relevant information needed for terminal correctness prediction can be low-dimensional in principle. The same applies to TD-error prediction, since $\delta_t$ is also a scalar. Therefore, a small number of hidden neurons can be sufficient, in principle, to preserve most of the value and TD-error information.

\paragraph{Realization.}
The superposition hypothesis~\citep{superposition} offers a principled explanation for why low-dimensional reward-related information may be realized as a sparse set of neurons. Specifically, the hypothesis suggests that features that are frequent and high-importance during training tend to receive dedicated, sparse representations. Value and TD-error information is useful throughout reasoning, as they summarize whether a partial trajectory is likely to succeed and whether a newly generated step improves that trajectory. Therefore, the superposition hypothesis predicts that such information is likely to be encoded sparsely. To be clear, this argument does not prove that these neurons must be sparse. However, in the experiments that follow, we find that these neurons indeed occupy less than 1\% of the total neurons, consistent with the prediction of the superposition hypothesis.
\section{Sparse Value Neurons in Large Language Models}
\label{sec:value}
In this section, we first introduce value neurons, a sparse subset of neurons whose activations encode the model's expected value of the current state.

\subsection{Training a Value Probe}
\label{sec:value_probe}

Consider an autoregressive LLM $\mathcal{M}$. The initial state for generation is defined as $s_0$. During the generation process, the LLM produces tokens sequentially. At each step $t$, an action is sampled such that $a_t \sim \mathcal{M}(\cdot | s_t)$, and the subsequent state is updated as $s_{t+1} = s_t \oplus a_t$. Suppose the model generates a total of $T$ new tokens. Given that modern decoder-only LLMs consist of multiple Transformer blocks, each hidden layer produces a corresponding representation for every state $s_t$. We denote the hidden state at the $l$-th layer and $t$-th step as $h(s_t, l)$. Here, a \emph{neuron} refers to an individual coordinate of the LLM hidden state.

To extract the reward information embedded within these representations, we introduce a value probe $V$. Following \citep{zhu2025llmknowsestimatingllmperceived}, we employ a two-layer multi-layer perceptron (MLP) with ReLU activation. The input dimension matches the dimensionality of the LLM's $l$-th layer hidden states, while the output is a scalar representing the predicted reward.
After computing $h(s_t, l)$, we feed it into $V$ to obtain the value prediction output $V(h(s_t, l))$.
For brevity, we define $V_{l}(s_t) = V(h(s_t, l))$. The probe is optimized using temporal difference (TD) learning. Let $r(s_T)$ be the final binary reward received. Given a discount factor $\gamma$, the TD error $\delta_t$ is defined as:
\begin{equation}
\delta_t = 
\begin{cases} 
r(s_T) - V_{l}(s_T), & \text{if } t = T + 1 \\
\gamma V_{l}(s_{t}) - V_{l}(s_{t-1}), & \text{otherwise}.
\end{cases}
\end{equation}
We conduct a layer-wise analysis to investigate the characteristics of the value neurons. For a given layer $l$ and a training dataset $\mathcal{D_T}$, our objective for the value probe is to minimize the expected TD error over the distribution of generated sequences:
\begin{equation}
\mathcal{L}_{\mathrm{TD}}(l) = \mathbb{E}_{s_0 \sim \mathcal{D_T}, s_{t} \sim \mathcal{M}(\cdot | s_{t-1})} \left[ \sum_{t=1}^{T+1} \delta_t^2 \right].
\end{equation}
TD learning bootstraps each
state's value from its successor and enforces step-level
temporal consistency, whereas direct regression onto the
final reward can rely on static question-level correlates.

\subsection{Identifying Value Neurons}
\label{sec:value_id}

Once the value probe is trained, we evaluate its performance on a validation dataset $\mathcal{D}_V$. During evaluation,  we measure the predicted value $V_{l}(s_0)$ directly from the hidden state at the initial input position $s_0$. 
This allows us to detect whether the value neurons have already formed an assessment of the potential reward before any tokens of the answer are generated. 

Specifically, we evaluate the correlation between $V_{l}(s_0)$ and $r(s_T)$. Higher correlation indicates better predictive capability. We utilize $AUC(V_{l}(s_0), r(s_T))$, the Area Under the Receiver Operating Characteristic (AUC), as our metric.

To substantiate the existence of sparse value neurons, we must demonstrate a small subset of the hidden states maintains sufficient predictive power regarding the value, so we perform pruning experiments.
Assume the input dimensionality of our value probe is $N$. We introduce a pruning ratio $p$. We prune $pN$ of these input dimensions, feeding only the most significant $(1-p)N$ dimensions into the value probe.  To prune the network, we calculate the $L_1$ norm of the weights connecting the input dimensions to the neurons in the first hidden layer and prune the input dimensions that correspond to the smallest $L_1$ weight norms, constructing a pruned value probe $V_{l}^p$. 
The surviving probe input dimensions define the set of
value neurons at that pruning ratio.

We then plot the relationship between $AUC(V_{l}^p(s_0), r(s_T))$ and the pruning ratio $p$. At each pruning ratio, we resample the validation set with replacement 1,000 times and compute the standard deviation of the resulting AUC values to estimate uncertainty. If the AUC curve remains stable as $p$ increases, it provides strong evidence for the existence of value neurons.

\subsection{Empirical Evidence}

\begin{figure}[h]
    \centering
    \includegraphics[width=0.8\linewidth]{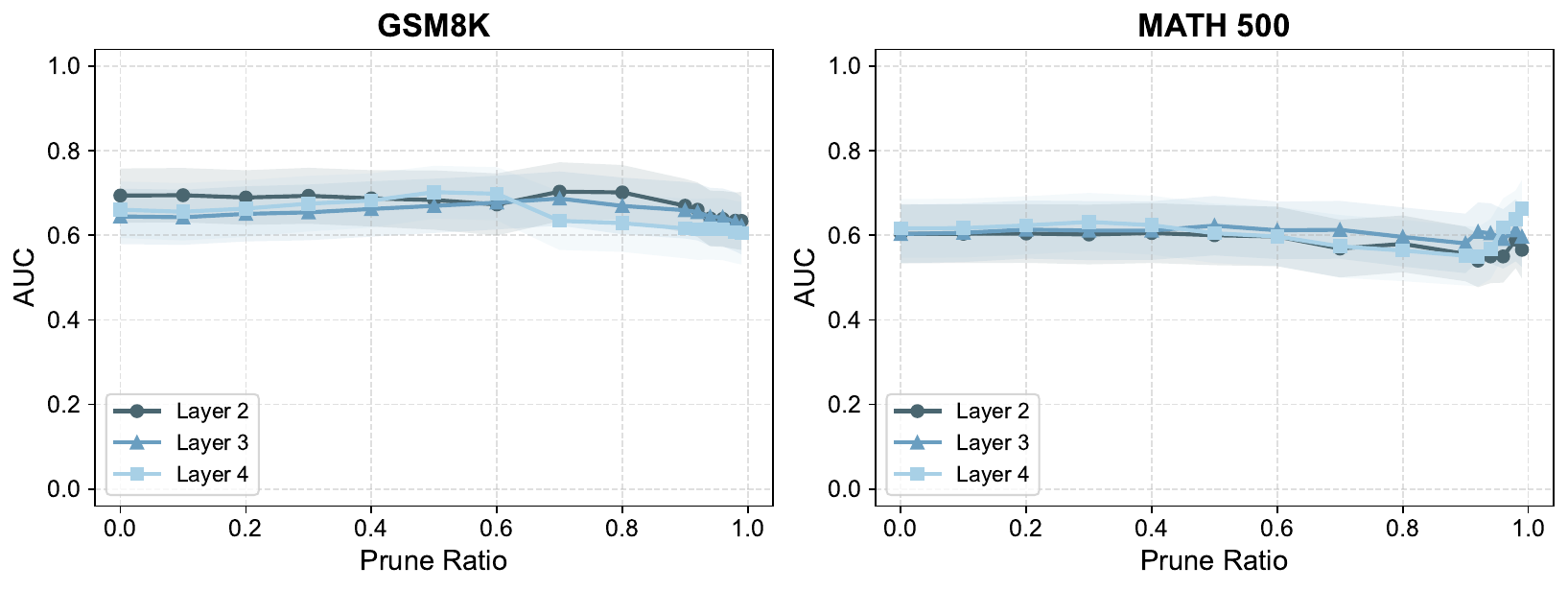}
    \caption{
        AUC curves for layers 2--4 of the Qwen-2.5-14B-SimpleRL-Zoo model on the GSM8K and MATH500 datasets. 
        The curves indicate that the value probe can accurately predict the value by relying on only a very small number of value neurons. Shaded regions indicate uncertainty.
    }
    \label{fig:value_neurons} 
    \vspace{-0.5em}
\end{figure}

We provide an illustrative example using the curves from layers 2--4 of the Qwen-2.5-14B-SimpleRL-Zoo ~\citep{zeng2025simplerlzooinvestigatingtamingzero} model on the GSM8K ~\citep{cobbe2021trainingverifierssolvemath} and MATH500 ~\citep{lightman2023letsverifystepstep} datasets. For detailed  hyperparameters, please refer to Appendix~\ref{app:hyper}. 
As illustrated in Figure \ref{fig:value_neurons}, the AUC curves do not exhibit a significant decline as pruning proceeds, which indicates that the value probe can effectively estimate the value of the current state by relying on a very small fraction (less than 1\%) of the total neurons. These specific neurons are designated as \textit{value neurons}.

\vspace{-0.5em}

In Appendix~\ref{app:rob}, we systematically verify the robustness of value neurons. Specifically, we validate the existence of sparse value neurons across benchmarks including MATH500~\citep{lightman2023letsverifystepstep}, ARC~\citep{Clark2018ThinkYH}, MBPP+~\citep{mbpp} (coding), and IFEval~\citep{ifeval} (general instruction-following), and across models including Qwen3.5-0.8B~\citep{qwen3}, Phi-3.5-mini-instruct~\citep{phi3}, Llama-3.1-8B-Instruct~\citep{llama3}, and Qwen-2.5-14B-SimpleRL-Zoo~\citep{zeng2025simplerlzooinvestigatingtamingzero}, confirming that value neurons consistently emerge at various layers.

We also note that the experiments in this section evaluate value neurons at the initial input position $s_0$, which is inherently challenging and results in moderate absolute AUC values. In Appendix~\ref{app:last_token}, we conduct analysis at the final response position $s_T$ and observe the AUC remains mostly stable above 0.8, confirming that value neurons encode reward information throughout the generation process.

\subsection{Intervention Experiments: Value Neurons might be  Reward-Related}
\label{sec:inter}

The probing results above show that value neuron activations \emph{correlate} with reward information. A natural follow-up question is whether these neurons might be reward-related. 

\textbf{Necessity.} We first examine whether value neurons might be necessary for reasoning. We select the Qwen-2.5-7B-SimpleRL-Zoo model, zero out the activations of the top 1\% of value neurons in specific layers, and then measure the resulting performance drop on the MATH500 dataset. We compare against four baselines: (1)~\textbf{Random}: randomly zeroing out the same proportion of neurons; (2)~\textbf{Weight Magnitude}: zeroing out the top 1\% of neurons with the largest absolute weights in the MLP \texttt{down\_proj} layer; (3)~\textbf{Wanda}~\citep{wanda}: zeroing out the top 1\% of neurons selected by Wanda; and (4)~\textbf{NTP Neurons}: training a probe with a reward-unrelated objective (next-token prediction) using the same architecture and pruning procedure, then zeroing out its top 1\% neurons. We provide additional comparisons in Appendix~\ref{app:intervention_controls}.

\begin{table}[h]
    \centering
    \renewcommand{\arraystretch}{1.15}
    \caption{Intervention results for the Qwen-2.5-7B-SimpleRL-Zoo model on the MATH500 dataset. Performance is measured by accuracy (\%) after zeroing out a 1\% subset of neurons in a single layer. The original accuracy is 75.2\%.}
    \label{tab:intervention_results}
    \vspace{6pt}
    \begin{tabular*}{\textwidth}{c @{\extracolsep{\fill}} ccccc}
    \hline
    \textbf{Layer} & \textbf{Value Neurons} & \textbf{Random} & \textbf{NTP Neurons} & \textbf{Weight Magnitude} & \textbf{Wanda} \\ \hline
    2 & 37.0 (-38.2) & 77.0 (+1.8) & 76.4 (+1.2) & 74.6 (-0.6) & 74.8 (-0.4) \\
    3 & 13.6 (-61.6) & 73.4 (-1.8) & 76.8 (+1.6) & 61.6 (-13.6) & 48.2 (-27.0) \\
    4 & 29.4 (-45.8) & 73.8 (-1.4) & 77.4 (+2.2) & 75.4 (+0.2) & 67.0 (-8.2) \\
    5 &  1.2 (-74.0) & 74.4 (-0.8) & 75.2 (+0.0) & 75.0 (-0.2) & 73.4 (-1.8) \\
    Avg & 20.3 (-54.9) & 74.6 (-0.6) & 76.4 (+1.2) & 71.6 (-3.6) & 65.8 (-9.4) \\ \hline
    \end{tabular*}
\end{table}

As shown in Table~\ref{tab:intervention_results}, zeroing out value neurons causes degradation. Additionally, zeroing out NTP neurons has no effect, and the overlap between value neurons and NTP neurons is only 0.7\% (random baseline: 0.5\%), indicating that the value neurons identified by the reward-trained probe occupy distinct positions from those identified by a reward-unrelated probe.These results suggest that value neurons might be necessary for reasoning.

\textbf{Specificity.} Having examined their necessity for reasoning, we next examine whether value neurons might be specifically reward-related.
We compare three reasoning benchmarks
(MATH500, GSM8K, and ARC) with two factual-recall benchmarks
(PopQA~\citep{popqa} and EntityQuestions~\citep{entityquestions}). We use the probe trained on
MATH and zero out the top 1\% of value neurons in layer 2
of Qwen-2.5-7B-SimpleRL-Zoo. 

\begin{table}[h]
    \centering
    \renewcommand{\arraystretch}{1.15}
    \vspace{-6pt}
    \caption{Comparison of reasoning and factual-recall
    performance after zeroing out the top 1\% of value
    neurons in layer 2 of Qwen-2.5-7B-SimpleRL-Zoo,
    using the probe trained on MATH.
    Performance is measured by accuracy (\%).
    $\Delta$ denotes the change in percentage points.}
    \label{tab:intervention_recall}
    \vspace{6pt}
    \begin{tabular*}{\textwidth}{
        l @{\extracolsep{\fill}} ccccc}
        \hline
        & \textbf{MATH500}
        & \textbf{GSM8K}
        & \textbf{ARC}
        & \textbf{PopQA}
        & \textbf{EntityQuestions} \\
        \hline
        Task type
            & Reasoning & Reasoning & Reasoning
            & Factual recall & Factual recall \\
        Baseline
            & 75.2 & 93.0 & 82.2 & 42.6 & 27.2 \\
        Value Neurons
            & 37.0 & 52.8 & 66.2 & 46.2 & 24.2 \\
        $\Delta$
            & $-38.2$ & $-40.2$ & $-16.0$
            & $+3.6$ & $-3.0$ \\
        \hline
    \end{tabular*}
\end{table}

As shown in
Table~\ref{tab:intervention_recall}, zeroing out value neurons
degrades performance on the reasoning
benchmarks, while leaving factual-recall performance largely
unaffected. This suggests the possibility of reward-related specificity of the identified neurons.

\textbf{Functional coherence.} We also show in Appendix~\ref{app:corr} that zeroing out value neurons in earlier layers causes dopamine neurons in subsequent layers to lose their TD error predicting abilities, while zeroing out the same number of random neurons has minimal effect. This further supports that these neurons might form a functionally coherent subsystem organized around reward signals.

In summary, the intervention experiments provide three lines of preliminary evidence: (1)~\emph{Necessity}: value neurons might be critical for reasoning performance. (2)~\emph{Specificity}: value neurons might be specifically associated with reward signals. (3)~\emph{Functional coherence}: disrupting value neurons in earlier layers might impair the reward subsystem in subsequent layers. 

We note that these intervention experiments are intended to provide preliminary insights rather than conclusive causal evidence. Potential confounds remain, including the stability of the selected neurons across probes trained on different rollouts and the overlap between the selected neurons and coordinates with large activation magnitudes in the residual stream. We leave a more thorough investigation of these factors to future work.

\subsection{Transferability of the Value Neurons Across Different Datasets and Models}
\label{sec:transfer}

\begin{figure}[h]
    \centering
    \includegraphics[width=0.8\linewidth]{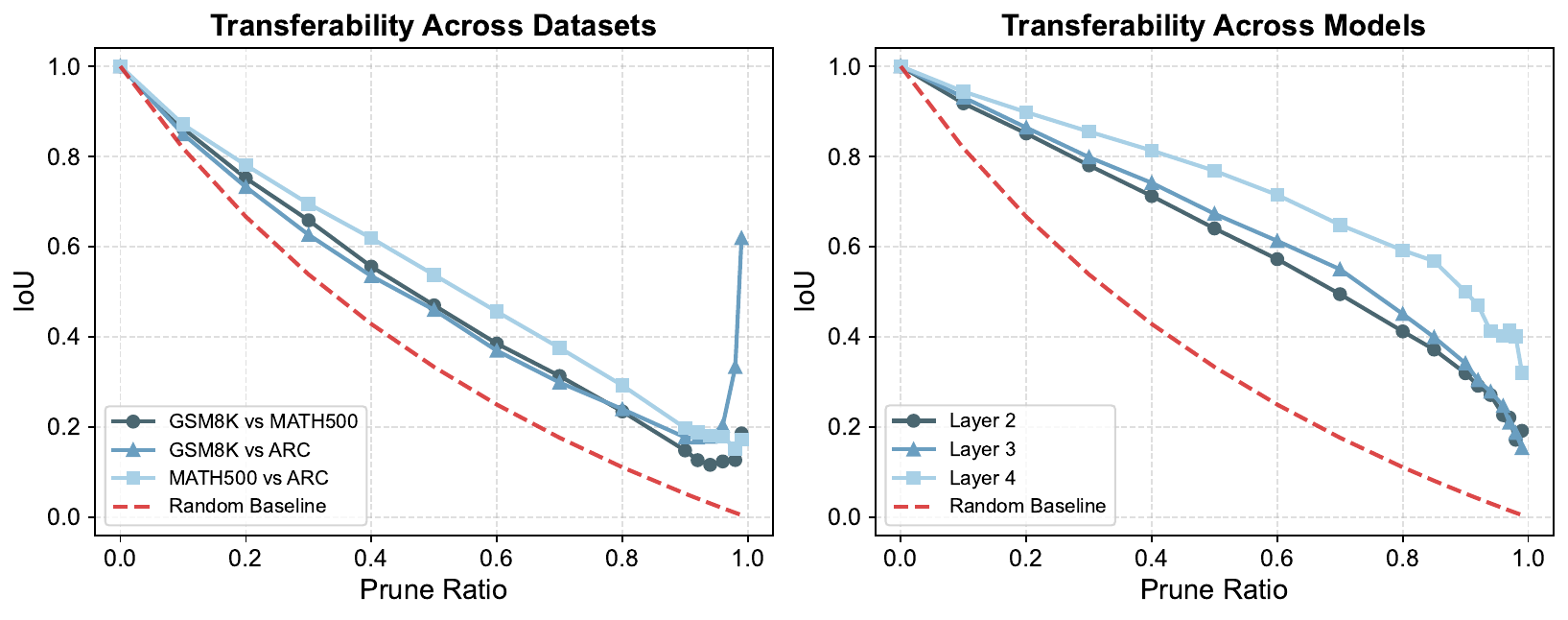}
    \caption{
        IoU as a function of the pruning ratio. The IoU values for value neurons across different datasets and models are significantly higher than the random baseline.
    }
    \label{fig:transfer} 
\end{figure}

If value neurons are an intrinsic property of LLMs, their positions should remain consistent across datasets and models. To test this, we compute the Intersection over Union (IoU) of value neurons identified on any two datasets at a given pruning ratio, and compare against a uniform selection baseline (derivation in Appendix~\ref{app:cal_random}). Using layer 3 of Qwen-2.5-14B-SimpleRL-Zoo, we compute pairwise IoU curves across three datasets (GSM8K, MATH500, ARC). As shown in Figure~\ref{fig:transfer}, IoU consistently exceeds the random baseline and increases sharply as the pruning ratio approaches 1, indicating that the most critical value neurons are highly stable across tasks. Additional results are provided in Appendix~\ref{app:iou}.

We further investigate whether value neuron positions are preserved across models fine-tuned from the same base model. We compare layers 2--4 of Qwen-2.5-7B-SimpleRL-Zoo and Qwen2.5-7B-PPO-Zero (both derived from Qwen-2.5-7B via RLVR), using value probes trained on MATH500. As shown in Figure~\ref{fig:transfer}, the IoU remains consistently above the random baseline, suggesting that value neurons are transferable across different models derived from the same base model.

\begin{figure*}[t]
    \centering
    \includegraphics[width=\textwidth]{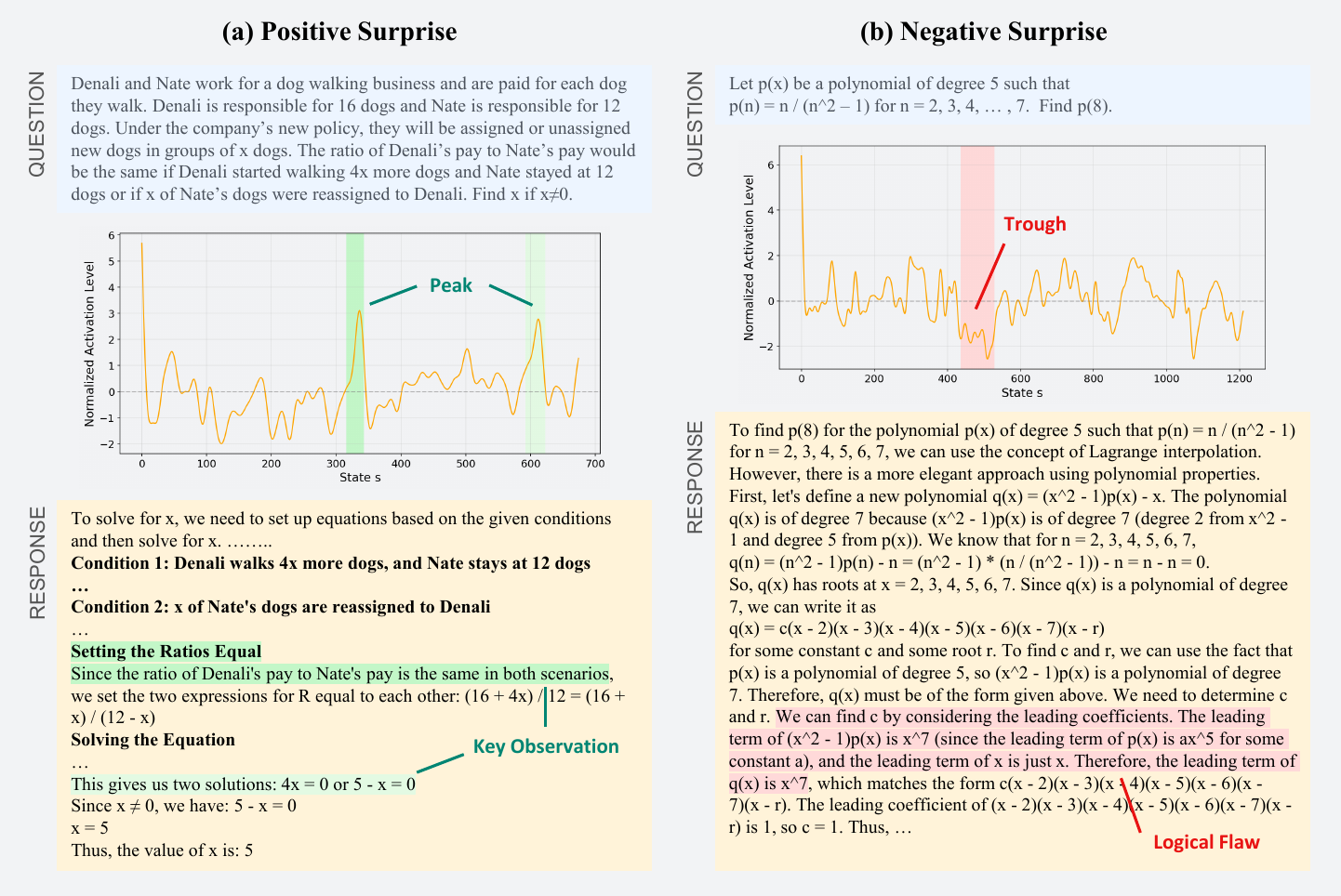}
    \caption{
       Dopamine neurons encode information regarding the model's prediction error for the current state.
(a) \textbf{Positive Surprise:} The model initially lacks confidence in answering the problem but ultimately provides the correct solution. This neuron exhibits two significant peaks when the model identifies a critical logical step and subsequently derives the final key result.
(b) \textbf{Negative Surprise:} Conversely, the model begins with high confidence but fails to solve the problem correctly. The neuron displays a distinct trough at the exact moment a logical flaw occurs.
    }
    \label{fig:dopamine_neurons_vis} 
    \vspace{-1.5em}
\end{figure*}

\section{Sparse Dopamine Neurons in Large Language Models}
\label{sec:dopamine}
In this section, we first introduce dopamine neurons, a sparse subset of neurons whose activations encode temporal-difference (TD) reward prediction errors. 

\subsection{Training a Dopamine Probe}
\label{sec:dopamine_probe}

 Given a problem, the model is initialized with state $s_0$ as defined in Section~\ref{sec:value}. The model generates a response $y$ which is segmented into $P$ paragraphs $y = (c_1, c_2, \ldots, c_P)$. The state at paragraph boundary $t$ is then defined as $   s_t = s_0 \oplus c_1 \oplus c_2 \oplus \cdots \oplus c_t$.
For each paragraph boundary, we estimate the value function $\hat{V}(s_t)$ by using Monte Carlo estimation to generate $K$ independent rollout completions from state $s_t$. At the terminal boundary, we set $\hat{V}(s_P) = r(s_P)$.
The TD-error at paragraph $t$ quantifies the unexpected change in reward:
\begin{equation}
    \delta_t = \gamma \, \hat{V}(s_t) - \hat{V}(s_{t-1}), \quad t = 1, 2, \ldots, P,
    \label{eq:td_error}
\end{equation}
where $\gamma \in [0, 1]$ is the discount factor. A positive $\delta_t$ indicates that paragraph $c_t$ increases the expected probability of arriving at a correct answer, while a negative $\delta_t$ indicates a detrimental reasoning step.

To extract the RPE information encoded in hidden states, we introduce a dopamine probe $D$, parallel in architecture to the value probe.
For paragraph $t$, we define $\tilde{h}(s_t, l)$ as the normalized hidden state in layer $l$ (see Appendix~\ref{app:detailed_dopamine} for details). The probe prediction is $D_l(s_t) = D(\tilde{h}(s_t, l))$. The probe is trained to minimize:
\begin{equation}
    \mathcal{L}_{\text{dopamine}}(l) = \mathbb{E}_{s_0 \sim \mathcal{D}_T} \left[ \sum_{t=1}^{P} \left( D_l(s_t) - \delta_t \right)^2 \right].
\end{equation}
To focus on steps where meaningful reward changes occur, we retain only paragraphs satisfying $|\delta_t| > 0.3$ during training. The same threshold is applied during evaluation. We provide an ablation study supporting the choice of this
threshold in Appendix~\ref{app:threshold_ablation}.

\subsection{Identifying Dopamine Neurons}
\label{sec:dopamine_id}

Once trained, we identify dopamine neurons via the same L1-norm pruning procedure used for value neurons. We introduce a pruning ratio $p$ and construct a pruned probe $D_l^p$. 
We evaluate the pruned probe on the validation set $\mathcal{D}_V$ via the Spearman correlation coefficient  $r\!\left(D_l^p(s_t),\, \delta_t\right)$ between predicted and MC-estimated TD errors.
If this metric remains stable as $p$ increases, it provides strong evidence that a small subset of neurons is sufficient to encode step-level reward prediction errors.

\subsection{Empirical Evidence}

\begin{figure}[h]
\centering
\includegraphics[width=0.8\linewidth]{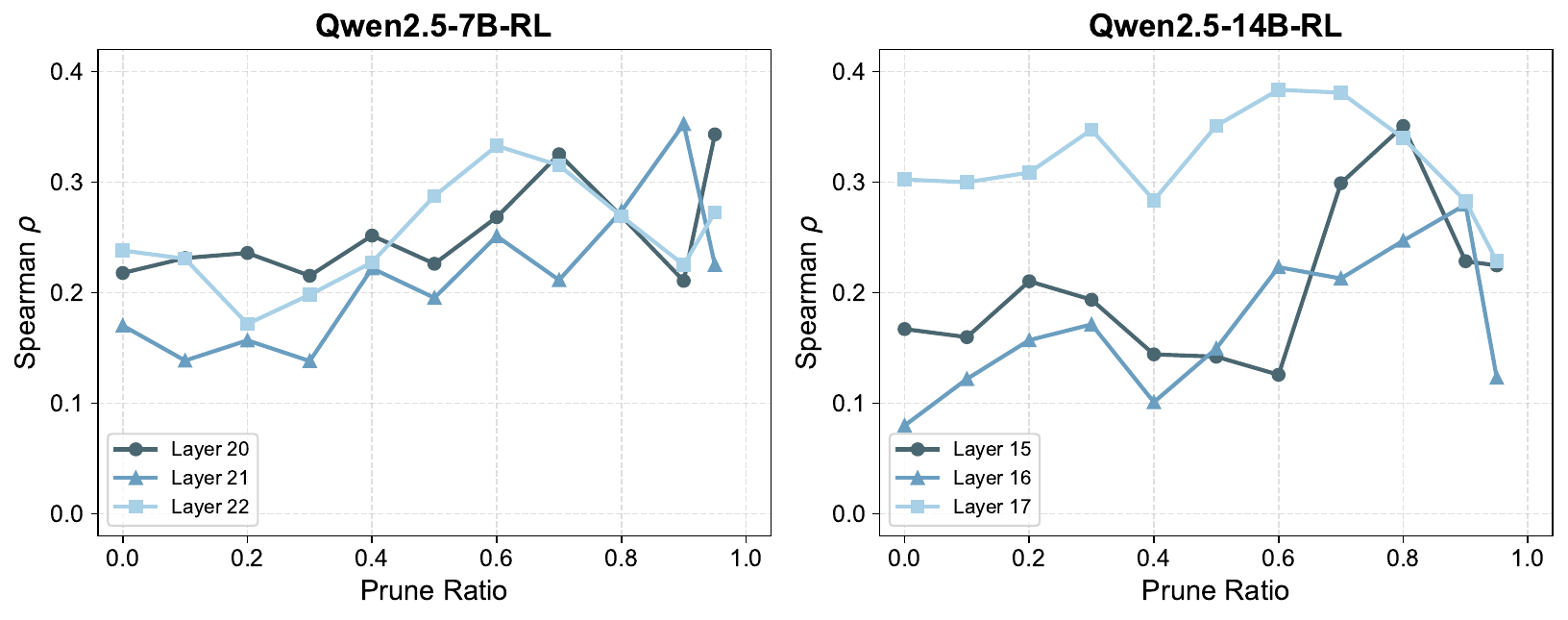}
\caption{Spearman correlation curves for Qwen-2.5-7B/14B-SimpleRL-Zoo  on MATH500.}
\label{fig:dopamine_neurons}
\end{figure}

Our experiments utilize the Qwen-2.5-7B/14B-SimpleRL-Zoo models. The probes are trained and evaluated on the MATH500 dataset. Please refer to Appendix~\ref{app:hyper} for details.

As illustrated in Figure \ref{fig:dopamine_neurons}, the Spearman correlation curves remain largely invariant across a wide range of pruning ratios. This demonstrates that the information necessary for signaling reward prediction errors is concentrated within a set of sparse neurons. We identify these as \textit{dopamine neurons}, as they encode reward prediction errors. 

\subsection{Visualization Results}

We illustrate the behavior of dopamine neurons using the $1517$-th neuron in layer 5 as an example. In Appendix~\ref{app:more_datasets}, we provide additional visualization cases.
As shown in Figure~\ref{fig:dopamine_neurons_vis}, in the positive surprise case (a), the model initially lacks confidence but ultimately solves the problem correctly. The neuron exhibits sharp activation peaks at the moments when the model identifies critical logical steps. In the negative surprise case (b), the model begins with high confidence but commits an error mid-reasoning, and the neuron displays a trough at the exact moment the logical flaw occurs. These patterns confirm that dopamine neuron activations closely track TD errors during inference.

\section{Applications}

\subsection{Applications of Value Neurons: Predicting Model Confidence}
\label{sec:app_confidence}
Since value neurons encode value information before any response tokens are generated, they can serve as a lightweight confidence estimator. This is practically valuable. For instance, modern reasoning models produce very long responses, so gauging confidence beforehand enables adaptive strategies such as dynamically allocating compute based on the initial confidence level.

We compare against four baselines across four models (Llama-3.1-8B, Phi-3.5-mini, Gemma-3-4B, Qwen3.5-0.8B) and two benchmarks (MATH500, ARC): LCD~\citep{cencerrado2025answerneededpredictingllm}, which trains a linear probe on the full hidden state; Verbal Confidence, which prompts the model to output its confidence directly; Token Confidence, which uses the next-token log probability at the question position; and Question Length as a simple proxy. For our method, AUC values are obtained at the optimal pruning ratio and averaged across layers 2--4. Baseline configurations are detailed in Appendix~\ref{app:baselines}.

As shown in Table~\ref{tab:confidence_full}, value neurons achieve the highest average AUC. This demonstrates that a sparse set of value neurons carries informative model confidence signals.

\begin{table*}[h]
\centering
\renewcommand{\arraystretch}{1.15}
\caption{Comparison of AUC for different confidence prediction methods across models and benchmarks. Higher is better. \textbf{Bold} denotes the best performing method per model.}
\begin{tabular*}{\textwidth}{l @{\extracolsep{\fill}} ccccc}
\toprule
\textbf{Model}
& \textbf{Value Neurons}
& \textbf{LCD~\citep{cencerrado2025answerneededpredictingllm}}
& \textbf{Verbal Conf.}
& \textbf{Token Conf.}
& \textbf{Question Len.} \\
\midrule
\multicolumn{6}{l}{\textbf{\textit{MATH500}}} \\
Llama-3.1-8B   & \textbf{0.69} & 0.64 & \textbf{0.69} & 0.54 & 0.63 \\
Phi-3.5-mini   & \textbf{0.73} & 0.63 & 0.54 & 0.41 & 0.62 \\
Gemma-3-4B     & \textbf{0.72} & 0.65 & 0.62 & 0.41 & 0.71 \\
Qwen3.5-0.8B   & 0.71 & 0.70 & 0.41 & 0.56 & \textbf{0.77} \\
\midrule
\multicolumn{6}{l}{\textbf{\textit{ARC}}} \\
Llama-3.1-8B   & \textbf{0.66} & 0.56 & 0.55 & 0.49 & 0.55 \\
Phi-3.5-mini   & \textbf{0.63} & 0.54 & 0.29 & 0.46 & 0.54 \\
Gemma-3-4B     & \textbf{0.59} & 0.50 & 0.47 & 0.44 & 0.49 \\
Qwen3.5-0.8B   & 0.63 & 0.54 & 0.55 & 0.50 & \textbf{0.65} \\
\midrule
\textbf{Average} & \textbf{0.67} & 0.60 & 0.52 & 0.48 & 0.62 \\
\bottomrule
\end{tabular*}
\label{tab:confidence_full}
\vspace{-1em}
\end{table*}

\subsection{Applications of Dopamine Neurons: Process Reward Model}
\label{sec:app_prm}
Since dopamine neurons encode step-level reward prediction errors, their activations 
can be directly leveraged to score individual reasoning steps, functioning as a 
Process Reward Model (PRM). To evaluate this, we employ a guided search procedure: at each paragraph boundary, the model generates $K=4$ candidate responses, each is scored by the PRM, and the highest-scoring candidate is selected as the next reasoning step. This procedure repeats until the model produces a final answer.

\begin{table}[h]
    \centering
    \caption{Comparison of accuracy (\%) on MATH500 for
    different methods.}
    \label{tab:prm_results}
    \setlength{\tabcolsep}{3pt}
    \renewcommand{\arraystretch}{1.15}
    \begin{tabularx}{\textwidth}{
        @{}lcc>{\centering\arraybackslash}Xc@{\hspace{6pt}}}
        \toprule
        \multicolumn{1}{c}{\makecell{\textbf{Method}}}
        & \makecell{Qwen-2.5-7B\\SimpleRL-Zoo}
        & \makecell{Qwen-2.5-14B\\SimpleRL-Zoo}
        & \makecell{Llama-3.1-8B Instruct}
        & \makecell{\textbf{Average}} \\
        \midrule
        Greedy
            & 69.6 & 67.0 & 47.3 & 61.3 \\
        Random
            & 77.3 & 74.8 & 46.3 & 66.1 \\
        Majority of N
            & 64.8 & 76.0 & 54.2 & 65.0 \\
        Implicit PRM
            & 69.8 & 74.8 & 45.8 & 63.5 \\
        Mean logprob
            & 72.2 & 78.7 & 41.7 & 64.2 \\
        Min logprob
            & 74.1 & 76.0 & 48.6 & 66.2 \\
        P(true)
            & 78.7 & 76.0 & 47.2 & 67.3 \\
        Dopamine Neurons
            & \textbf{79.6} & \textbf{82.7}
            & \textbf{54.6} & \textbf{72.3} \\
        \bottomrule
    \end{tabularx}
\end{table}

We compare the dopamine probe against seven baselines on MATH500 using three models.
\textbf{Greedy} decoding uses $K=1$ with temperature zero.
\textbf{Random} selects randomly among the $K$ candidates.
For \textbf{Implicit PRM}, motivated by the theoretical result that   KL-constrained RL yields $\pi^*\!\propto\pi_{\mathrm{ref}}\exp(r/\beta)$  \citep{rafailov2024from,yuan2025free}, we use  $\log\pi_{\mathrm{RL}}-\log\pi_{\mathrm{base}}$ as an approximation to the step-level reward. 
\textbf{Majority of N} selects the most frequent final answer
among $N$ sampled responses;
\textbf{Mean logprob}~\citep{meanlogp} and \textbf{Min logprob}~\citep{minlogp} score
candidates using the mean and minimum token log probabilities,
respectively; and
\textbf{P(true)}~\citep{kadavath2022languagemodelsmostlyknow} uses the model's self-estimated probability
that a candidate is correct.
In our method, we use the predicted value of the dopamine probe trained on the optimal layer as the process reward. All experiments are averaged over three random seeds.

As shown in Table~\ref{tab:prm_results}, using the dopamine neuron predictions as a process reward signal improves the model's reasoning performance, demonstrating dopamine neurons can be applied to guide reasoning.

\section{Related Work}
\paragraph{Probing methods in large language models.}
Probing methods serve as powerful tools for investigating and interpreting the internal characteristics of Large Language Models (LLMs) and have been extensively utilized within the research community~\citep{languagemodelcircuits,harp,factcheckmate,gurnee2023finding}. No-answer-needed~\citep{cencerrado2025answerneededpredictingllm} demonstrated that linear probes can be trained to predict the correctness of a model's forthcoming answer.  Expanding beyond linear analysis, \citep{diegosimón2024polarcoordinaterepresentssyntax} introduced a polar probe designed to extract syntactic relations by analyzing both the distance and direction between word embeddings.
Probing has also been instrumental in assessing the veracity of generated content. \citep{marks2024the} found that LLMs linearly represent the truth or falsehood of factual statements. This is supported by \citep{han-etal-2025-simple}, who noted that LLM hidden states are highly predictive of factuality in long-form natural language generation, and that such information can be efficiently extracted at inference time using a lightweight probe. \citep{liang2025neuralprobebasedhallucinationdetection} employed lightweight MLP probes to perform nonlinear modeling of high-level hidden states for token-level hallucination detection. \citep{gao2025hneuronsexistenceimpactorigin} identified that a  sparse subset of neurons within LLMs can reliably predict the occurrence of hallucinations.
Furthermore, \citep{heindrich2025do} observed that simple probing methods demonstrate superior generalization on Out-of-Distribution (OOD) tasks compared to Sparse Autoencoders (SAEs). Following  \citep{zhu2025llmknowsestimatingllmperceived}, our study employs a  two-layer MLP network as our probing model to maintain simplicity while capturing the necessary reward signals.

\paragraph{Reward modeling based on LLM internal representations.}
Recent studies have investigated the extraction of reward signals from the internal representations of LLMs~\citep{li2023inferencetime}. Anthropic proposed a value head to predict whether models can answer questions correctly~\citep{kadavath2022languagemodelsmostlyknow}. \citep{burns2023discovering} utilized a purely unsupervised approach to discover latent knowledge within the internal activations of a language model, enabling the accurate answering of yes-no questions using only model activations. 
% \citep{Hajrullahu2025} identified a significant positive correlation between the statistics of a model's hidden layers and the correctness of its final output, noting that larger models exhibit internal states with stronger predictive power regarding rewards. 
\citep{zhang2025interpretablerewardmodelsparse} utilized tools from mechanistic interpretability to analyze model activations, employing Sparse Autoencoders (SAEs) to map high-dimensional hidden states onto interpretable semantic features. Similarly, the RISE framework~\citep{liu2025trust}  proposed simultaneously enhancing a model's problem-solving and self-verification capabilities within a single training process, requiring the model to generate both a solution and a corresponding evaluative score.
Furthermore, \citep{zhao2025learningreasonexternalrewards} introduced Reinforcement Learning from Internal Feedback (RLIF), a framework that enables LLMs to learn from intrinsic signals in the absence of external rewards.  Similarly, LaSeR~\citep{yang2025laserreinforcementlearninglasttoken} found that a last-token self-rewarding score can  guide the reinforcement learning process. \citep{latentthinking} observed that the latent thoughts leading to correct versus incorrect answers exhibit highly distinguishable patterns, allowing a latent classifier to   predict answer correctness directly from these representations.  The SWIFT method~\citep{guo2025mining} demonstrated that mining intrinsic rewards from LLM hidden states facilitates efficient Best-of-N sampling.
Despite these attempts to predict rewards from internal representations, existing literature has yet to identify how this information is structured within the hidden states.
\section{Conclusion}
In this work, we show that reward information in LLM hidden states is concentrated in a sparse set of neurons, the reward subsystem. Using probing and pruning, we identify value neurons that encode value information and dopamine neurons that encode reward prediction errors. Intervention experiments confirm that value neurons are specifically tied to reward signals, and their positions transfer across datasets and models. On the application side, value neurons provide lightweight confidence estimates, and dopamine neurons can serve as an intrinsic process reward model. Looking ahead, we believe the reward subsystem offers a useful lens for understanding LLM reasoning, and we hope this work encourages further exploration of neuron-level structure in language models.

\clearpage
\bibliography{example_paper}
\bibliographystyle{plainnat}

%%%%%%%%%%%%%%%%%%%%%%%%%%%%%%%%%%%%%%%%%%%%%%%%%%%%%%%%%%%%

\clearpage
\appendix

\etocdepthtag.toc{appendix}
\etocsettocstyle{\section*{Appendix - Table of Contents}}{}
\tableofcontents
\vspace{1em}

\onecolumn
\section{Hyperparameters and Experimental Settings}
\label{app:hyper}
Response generation is performed on two NVIDIA RTX PRO 6000 Blackwell Server Edition GPUs. Both training and inference are conducted on a single NVIDIA RTX PRO 6000 Blackwell Server Edition GPU.

\subsection{Response Generation Hyperparameters}
The hyperparameters used for response generation are summarized in Table~\ref{tab:hyperparams_generation}.

\begin{table*}[h]
\centering
\caption{Hyperparameters used for response generation.}
\label{tab:hyperparams_generation}

\begin{tabularx}{\textwidth}{@{}lX@{}}
\toprule
\textbf{Category} & \textbf{Hyperparameter = Value} \\
\midrule
\textbf{Generation} &
max\_tokens\_per\_call$\ =\ $16000; temperature$\ =\ $1.0; \\
& top\_p$\ =\ $0.95; n\_sampling$\ =\ $1; prompt\_type$\ =\ $qwen-boxed \\
\midrule
\textbf{Inference} &
use\_vllm$\ =\ $True; gpu\_memory\_utilization$\ =\ $0.75; \\
\bottomrule
\end{tabularx}

\end{table*}

\subsection{Value / Dopamine Probe Training Hyperparameters}
The hyperparameters used for value and dopamine probe training are listed in Table~\ref{tab:hyperparams_training}.

\begin{table*}[h]
\centering
\caption{Hyperparameters used for value and dopamine probe training.}
\label{tab:hyperparams_training}

\begin{tabularx}{\textwidth}{@{}lX@{}}
\toprule
\textbf{Category} & \textbf{Hyperparameter = Value} \\
\midrule
\textbf{Optimization} &
optimizer$\ =\ $AdamW; learning\_rate$\ =\ $1e-4;  weight\_decay$\ =\ $0.01 \\
\midrule
\textbf{Training} &
num\_epochs$\ =\ $100; batch\_size$\ =\ $4; train\_ratio$\ =\ $0.8; gamma$\ =\ $1$-$1e-5; seed$\ =\ $0\\
\midrule
\textbf{Architecture} &
hidden\_size$\ \rightarrow\ $1024 (value) / 32 (dopamine)$\ \rightarrow\ $1 (two-layer MLP) \\
\bottomrule
\end{tabularx}

\end{table*}

\subsection{AUC Curve Evaluation Hyperparameters}
The hyperparameters used for reward subsystem evaluation with neuron pruning are provided in Table~\ref{tab:hyperparams_evaluation}.

\begin{table*}[h!]
\centering
\caption{Hyperparameters used for reward subsystem evaluation with neuron pruning.}
\label{tab:hyperparams_evaluation}
\begin{tabularx}{\textwidth}{@{}lX@{}}
\toprule
\textbf{Category} & \textbf{Hyperparameter = Value} \\
\midrule
\textbf{Evaluation} &
batch\_size$\ =\ $4; train\_ratio$\ =\ $0.8; seed$\ =\ $0\\
\midrule
\textbf{Pruning} &
prune\_ratios$\ =\ $\{0.0, 0.1, 0.2, 0.3, 0.4, 0.5, 0.6, 0.7, 0.8, 0.9, 0.92, 0.94, 0.96, 0.98, 0.99\}; \\
\bottomrule
\end{tabularx}

\end{table*}

\section{Derivation of the IoU Curve for the Random Baseline}
\label{app:cal_random}
 Let $N$ denote the total number of neurons and $p \in [0,1)$ denote the pruning ratio. Each set contains $k = (1-p) N$ neurons selected uniformly at random. For two independently sampled sets $A$ and $B$, each of size $k$, the intersection size $|A \cap B|$ follows a hypergeometric distribution:
$$P(|A \cap B| = i) = \frac{\binom{k}{i} \binom{N-k}{k-i}}{\binom{N}{k}}$$
where $i \in [\max(0, 2k-N), k]$. The expected IoU is given by:

$$\mathbb{E}\left[\text{IoU}\right] =  \sum_{i=\max(0,2k-N)}^{k} \frac{\binom{k}{i} \binom{N-k}{k-i}}{\binom{N}{k}} \cdot \frac{i}{2k - i}$$

\section{Robustness of the Value Neurons}
\label{app:rob}
 In this section, we demonstrate that value neurons are robust across various datasets, model scales, layers, and model architectures. 

\subsection{Robustness across Different Datasets}
To demonstrate the robustness of value neurons, we investigate whether these findings generalize to other representative task types, such as coding and general instruction-following. To this end, we extend our verification to include MBPP+~\citep{mbpp} (coding) and IFEval~\citep{ifeval} (general instruction-following).

\begin{figure}[h]
    \centering
    \includegraphics[width=0.8\linewidth]{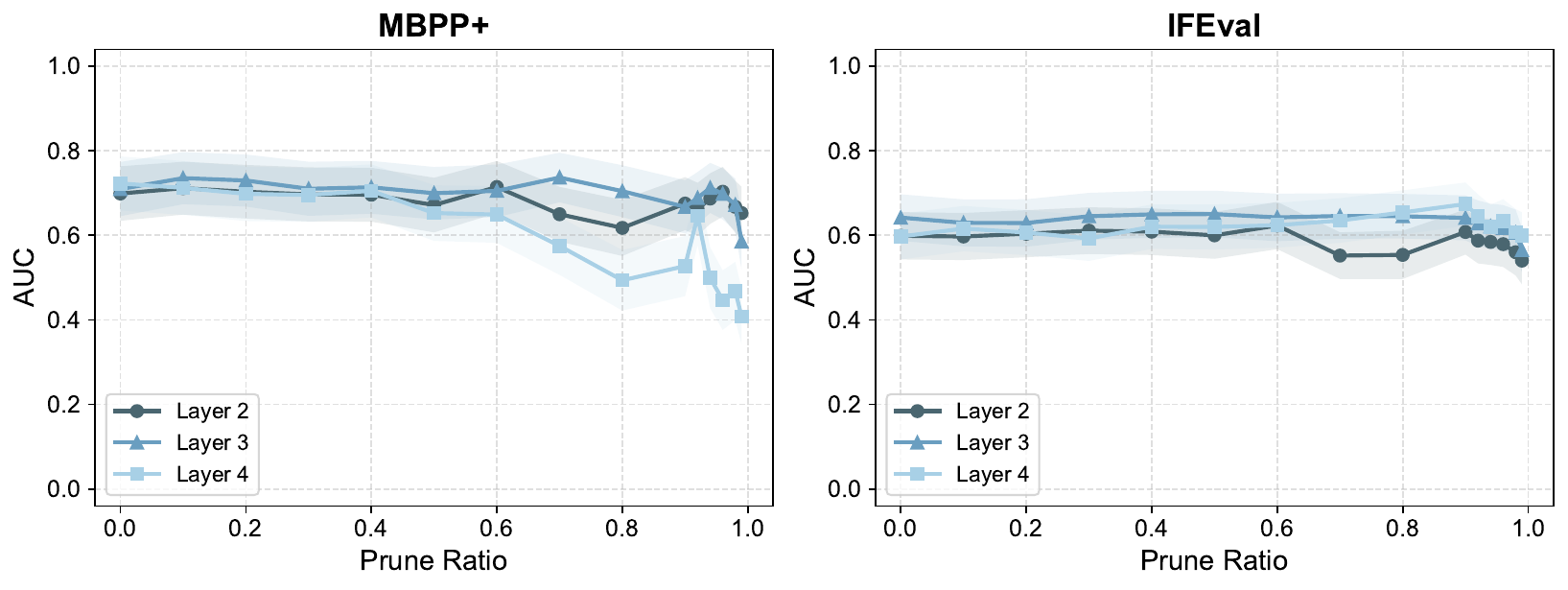}
    \caption{
        AUC curves for layers 2--4 of the Qwen-2.5-7B-SimpleRL-Zoo model on the MBPP+ and IFEval datasets. Shaded regions indicate uncertainty.
    }
    \label{fig:more_datasets} 
\end{figure}

 As shown in Figure~\ref{fig:more_datasets}, the AUC curve remains largely invariant to pruning, demonstrating that the existence of value neurons within the reward subsystem is consistently observed across different types of datasets.

\subsection{Robustness across Different Models}

We primarily focus on models based on the Qwen architecture. In addition to Qwen~\citep{qwen}, the Llama~\citep{llama}  and Phi~\citep{phi} architectures are also widely adopted in the open-source LLM community. To verify the applicability of the reward subsystem across these architectures, we conduct experiments on the MATH500 and ARC dataset using Qwen3.5-0.8B~\citep{qwen3}, Phi-3.5-mini-instruct~\citep{phi3}, Llama-3.1-8B-Instruct~\citep{llama3}, and Qwen-2.5-14B-SimpleRL-Zoo models.

\begin{figure}[h]
    \centering
    \includegraphics[width=\linewidth]{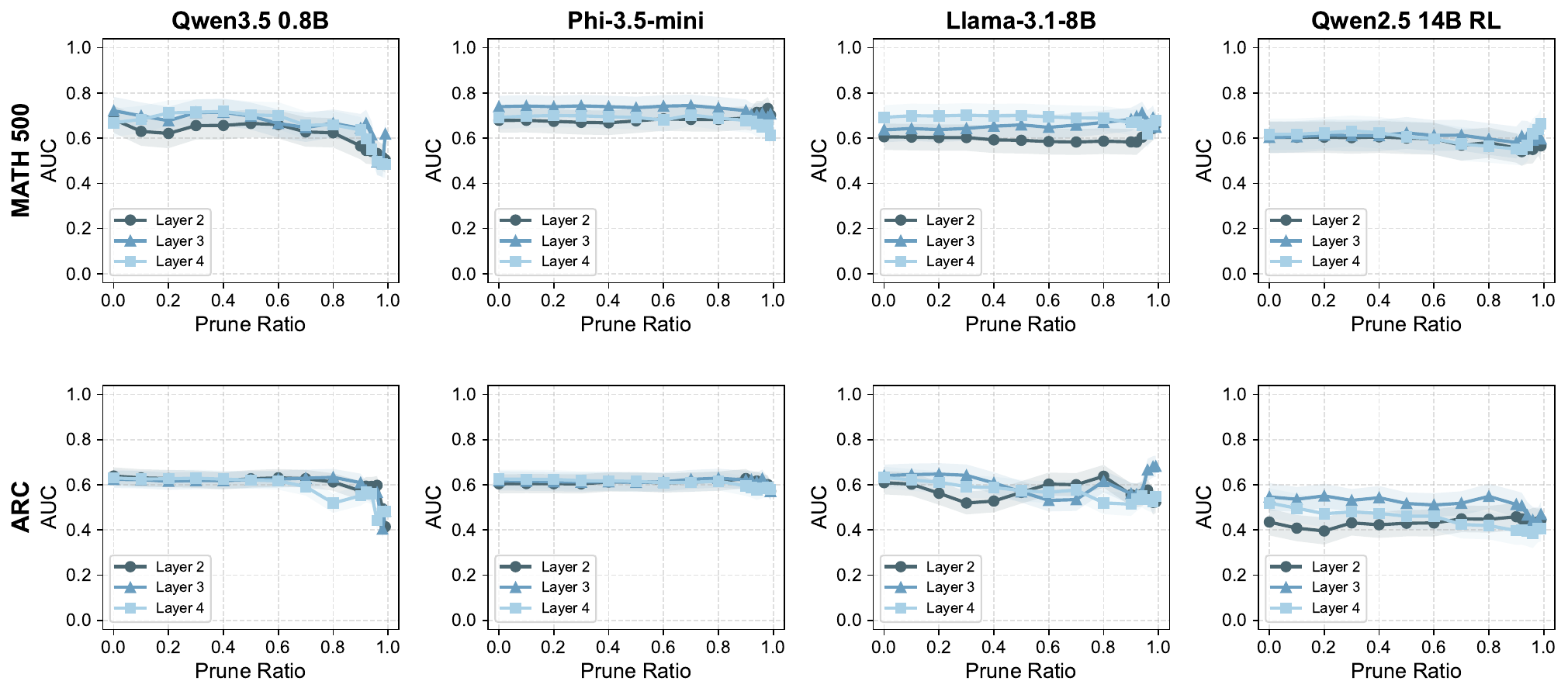}
    \caption{
        AUC curves for layers 2--4 of the Qwen-3.5-0.8B, Phi-3.5-mini-instruct, Llama-3.1-8B-Instruct, and Qwen-2.5-14B-SimpleRL-Zoo models on the MATH500 and ARC dataset. Shaded regions indicate uncertainty.
    }
    \label{fig:more_models} 
\end{figure}

As illustrated in Figure~\ref{fig:more_models}, our conclusion regarding the existence of sparse value neurons remains valid for open-source models with different architectures and model scales.

\subsection{Robustness across Different Layers}
 In this section, we examine the robustness of the reward subsystem across various layers. For this purpose, we utilize the Qwen-2.5-7B/14B-SimpleRL-Zoo model.

\begin{figure}[h]
    \centering
    \includegraphics[width=0.8\linewidth]{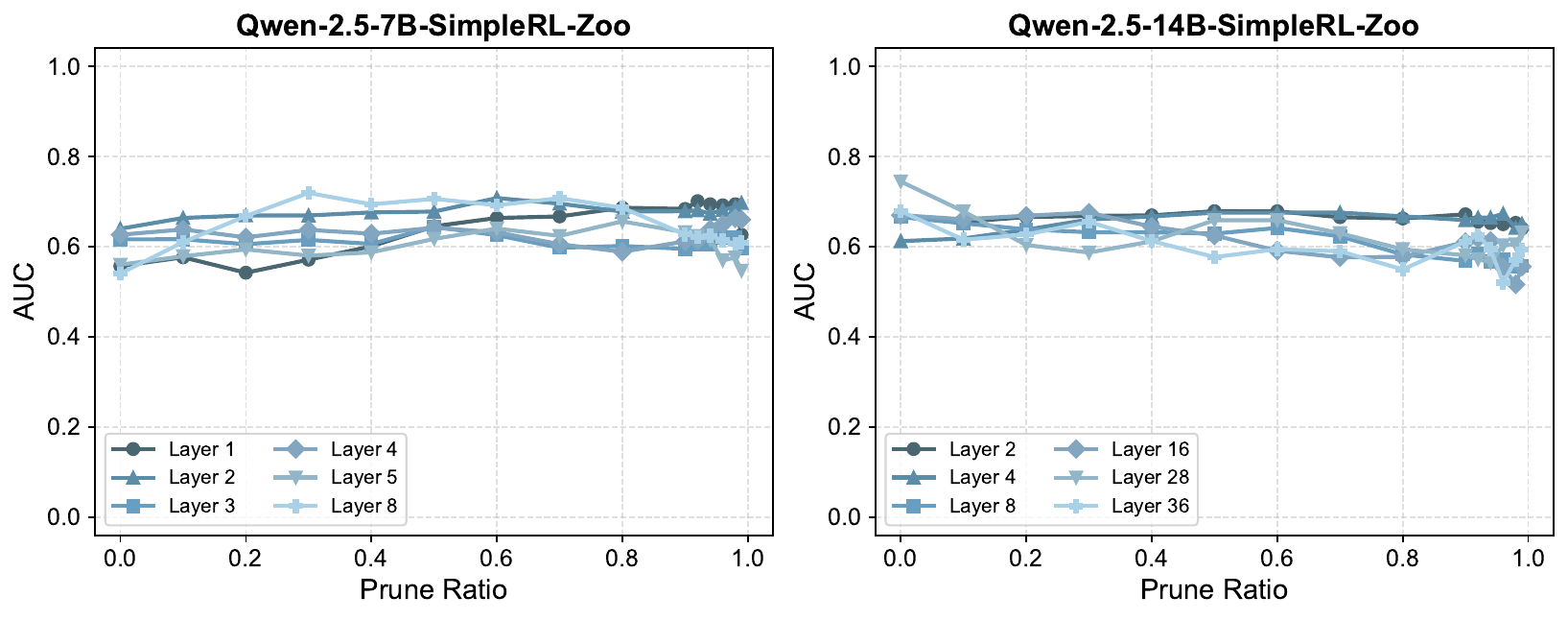}
    \caption{
        AUC curves for different layers of the Qwen-2.5-7B/14B-SimpleRL-Zoo model on the GSM8K dataset. 
    }
    \label{fig:more_layers} 
\end{figure}

As shown in Figure~\ref{fig:more_layers}, we find that the AUC remains largely stable as the pruning ratio rises.

\section{Transferability Across Different Datasets: More IoU Curves}
\label{app:iou}

\begin{figure}[h]
    \centering
    \includegraphics[width=0.8\linewidth]{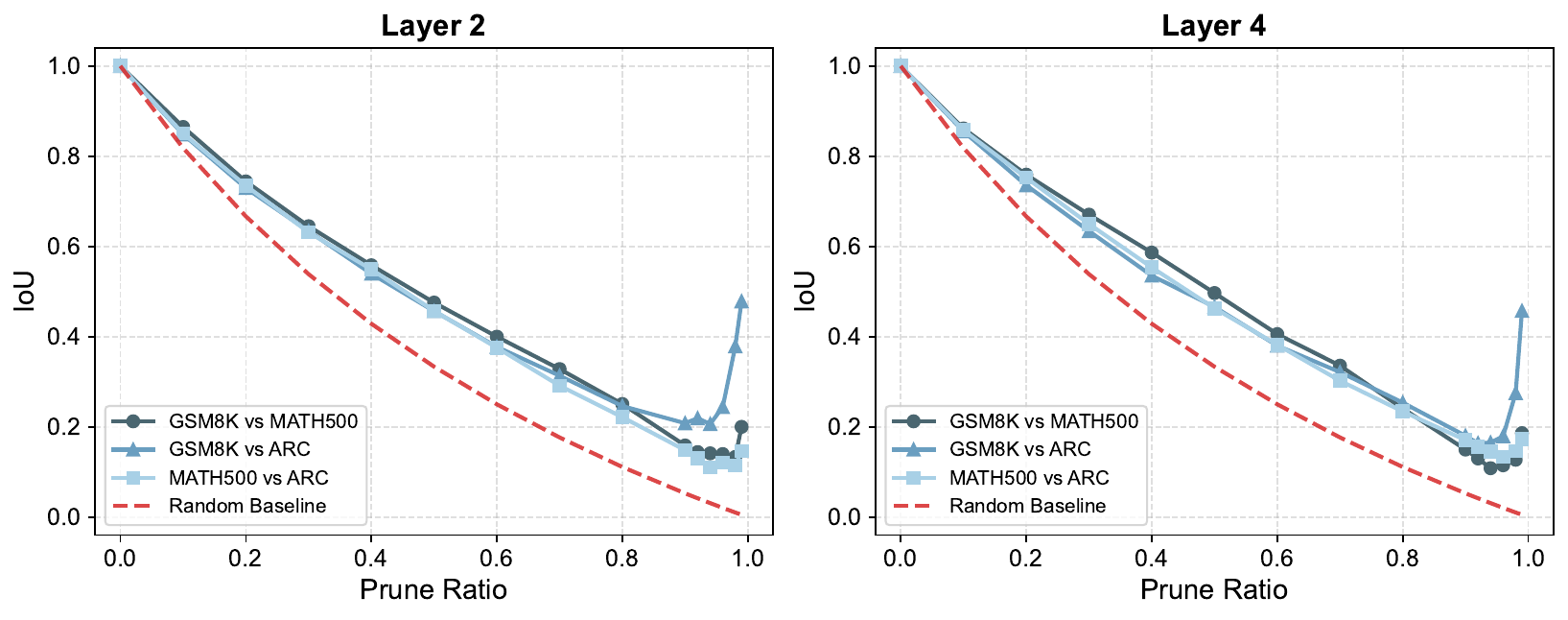}
    \caption{
        IoU as a function of the pruning ratio. The IoU values for value neurons across different datasets are significantly higher than the random baseline, indicating that for the same LLM, the positions of identified value neurons are closely correlated across tasks.
    }
    \label{fig:transfer_dataset_1} 
\end{figure}

In this section, we provide additional evidence regarding the transferability of value neurons across different datasets. As illustrated in Figure~\ref{fig:transfer_dataset_1}, we utilize layer 2 and layer 4 of the Qwen-2.5-14B-SimpleRL-Zoo model and compute pairwise IoU curves as a function of the pruning ratio for the GSM8K, MATH500, and ARC datasets.
In these results, the IoU curves for any two datasets consistently exceed the random baseline. Furthermore, the observation that many IoU curves exhibit a significant upward trend as the pruning ratio approaches 1 remains valid for these layers as well.

\section{Detailed Procedures for Hidden State Normalization}
\label{app:detailed_dopamine}

For each sample, we perform a single forward pass through the full sequence and extract hidden states from all $L$ transformer layers. Let $h_{j,n,l}$ denote the hidden state at token position $j$ for the $n$-th neuron in layer $l$.

To remove sample-specific baseline activation levels, we apply $z$-score normalization across the token dimension within each response. Let $j_0$ denote the token index at which the response begins. For each neuron, we compute:
\begin{equation}
    \bar{h}_{n,l} = \frac{1}{|\mathcal{T}|}\sum_{j \in \mathcal{T}} h_{j,n,l}, \quad
    \sigma_{n,l} = \sqrt{\frac{1}{|\mathcal{T}|}\sum_{j \in \mathcal{T}} (h_{j,n,l} - \bar{h}_{n,l})^2},
\end{equation}
where $\mathcal{T} = \{j_0, j_0+1, \ldots, J\}$ is the set of response token positions and $J$ is the last token position. The normalized activation is:
\begin{equation}
    \tilde{h}_{j,n,l} = \frac{h_{j,n,l} - \bar{h}_{n,l}}{\sigma_{n,l} + \epsilon}, \quad \epsilon = 10^{-8}.
\end{equation}

Motivated by neuroscience, where dopaminergic responses are measured as average firing rates over a temporal window, we aggregate activations at the paragraph level rather than at individual token positions.
 Let $\mathcal{T}_t \subset \mathcal{T}$ denote the set of token positions corresponding to paragraph $c_t$. The paragraph-level representation is the mean of the normalized activations over its tokens:
\begin{equation}
    \tilde{h}(s_t, l) = \left[ \frac{1}{|\mathcal{T}_t|} \sum_{j \in \mathcal{T}_t} 
    \tilde{h}_{j,n,l} \right]_{n=1}^{d} \in \mathbb{R}^{d}.
\end{equation}

\section{Further Evidence for the Characteristics of Dopamine Neurons}
\label{app:more_datasets}

\begin{figure*}[h]
    \centering
    \includegraphics[width=\textwidth]{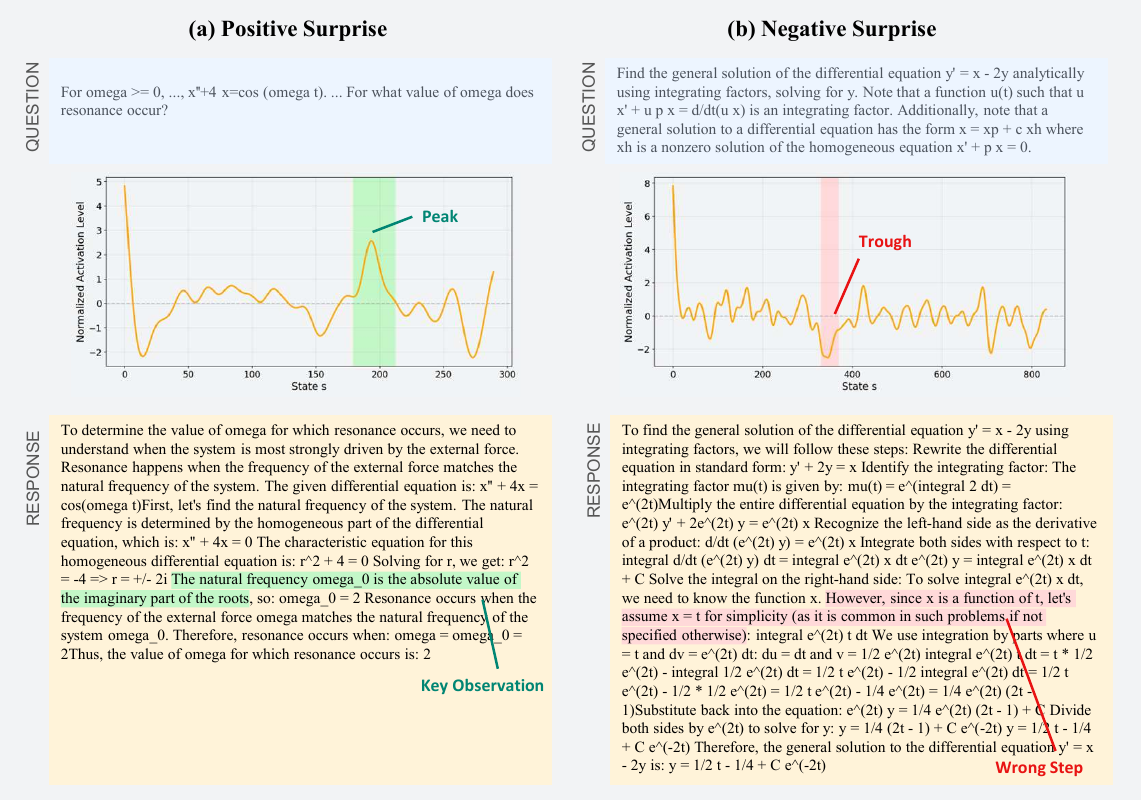}
    \caption{
        Dopamine neurons encode information regarding the model's prediction error for the current state.
        (a) \textbf{Positive Surprise:} The model initially lacks confidence in answering the problem but ultimately provides the correct solution. This neuron exhibits a significant peak when the model identifies a critical logical step and subsequently derives the final key result.
        (b) \textbf{Negative Surprise:} Conversely, the model begins with high confidence but fails to solve the problem correctly. The neuron displays a distinct trough at the exact moment a wrong step occurs.
    }
    \label{fig:more_datasets_app} 
\end{figure*}

To verify the robustness of the dopamine neurons' existence, we conduct further experiments using the Minerva Math dataset. We maintain the previously established experimental setup, selecting the top 50 neurons in each layer as dopamine neurons on the Minerva Math dataset. Notably, we observe a significant overlap between these dopamine neurons and those identified earlier on the MATH500 dataset; in particular, the $1517$-th neuron in layer 5 remains among the dopamine neurons. 

Consequently, we continue to investigate the predictive capacity of this neuron. Our findings confirm that it consistently displays a period of low activation when the model initially predicts a high value but fails to obtain a reward, and a period of high activation when the model initially predicts a low value but ultimately succeeds.

As illustrated in Figure~\ref{fig:more_datasets_app}, this neuron encounters both a positive surprise and a negative surprise. In the case of the positive surprise shown in Figure~\ref{fig:more_datasets_app}(a), the model initially exhibits low confidence in completing the task. However, during the inference process (around the 200th token), the model derives a key conclusion, resulting in a high TD error; consequently, we observe a sharp spike in the neuron's activation level. Since the subsequent reasoning proceeds steadily, the TD error remains low as the following steps become predictable, leading to the neuron's activation returning to a relatively low state. These observations suggest that dopamine neurons exhibit higher activation levels when the model acquires unexpected rewards or makes significant progress. 

Conversely, in the negative surprise shown in Figure~\ref{fig:more_datasets_app}(b), the model begins with high confidence. During the initial stage of inference, the model follows a correct path and even provides the critical modeling logic within the first 300 tokens. However, between the 300th and 400th tokens, the model commits an error, leading to a significant negative TD error and a corresponding suppression in the neuron's activation. 

Thus, our visualization on the Minerva Math dataset consistently demonstrates a close correlation between the TD error during inference and neuronal activation levels, further substantiating the fundamental properties of dopamine neurons.

\section{Additional Intervention Controls}
\label{app:intervention_controls}

We compare against four
additional baselines: \textbf{Massive Activations}~\citep{mass},
\textbf{Super Weights}~\citep{superw},
\textbf{Rogue Dimensions}~\citep{rogue}, and
\textbf{Attention Sink}~\citep{sink}. We conduct these experiments in
layer 2 of Qwen-2.5-7B-SimpleRL-Zoo on MATH500, GSM8K,
and ARC, using value neurons identified by the probe trained
on MATH.

\begin{table}[h]
    \centering
    \renewcommand{\arraystretch}{1.15}
    \caption{Additional intervention controls for layer 2 of
    Qwen-2.5-7B-SimpleRL-Zoo on MATH500, GSM8K, and ARC.
    Baseline denotes the unmodified model, and parentheses
    indicate changes in percentage points.}
    \label{tab:intervention_controls}
    \vspace{6pt}
    \begin{tabular*}{\textwidth}{
        l @{\extracolsep{\fill}} ccc}
        \hline
        \textbf{Method}
        & \textbf{MATH500}
        & \textbf{GSM8K}
        & \textbf{ARC} \\
        \hline
        Baseline
            & 75.2 & 93.0 & 82.2 \\
        Massive Activations
            & 76.0 (+0.8) & 91.6 (-1.4) & 83.0 (+0.8) \\
        Super Weights
            & 74.6 (-0.6) & 91.2 (-1.8) & 80.4 (-1.8) \\
        Rogue Dimensions
            & 65.8 (-9.4) & 81.6 (-11.4) & 79.8 (-2.4) \\
        Attention Sink
            & 73.2 (-2.0) & 93.0 (+0.0) & 81.6 (-0.6) \\
        \textbf{Value Neurons}
            & \textbf{37.0 (-38.2)}
            & \textbf{52.8 (-40.2)}
            & \textbf{66.2 (-16.0)} \\
        \hline
    \end{tabular*}
\end{table}

As shown in Table~\ref{tab:intervention_controls}, zeroing out
value neurons causes larger performance drops.

\section{Correlation between Value Neurons and Dopamine Neurons}
\label{app:corr}

Given that value neurons and dopamine neurons both belong to the reward subsystem, we hypothesize that they are functionally interconnected and exert mutual influence. To verify this, we perform an ablation experiment where we zero out the activation values of value neurons in earlier layers and observe the resulting changes in the activation trajectories of dopamine neurons in subsequent layers. As a control, we compare this with the effect of zeroing out an equivalent number of randomly selected neurons in the same layers. If perturbing the value neurons leads to a significantly more pronounced alteration in the dopamine neurons' activation curves compared to the random perturbation, it would indicate a close relationship between value neurons and dopamine neurons.

\begin{figure}[h]
    \centering
    \begin{minipage}{0.48\linewidth}
        \centering
        \includegraphics[width=\linewidth]{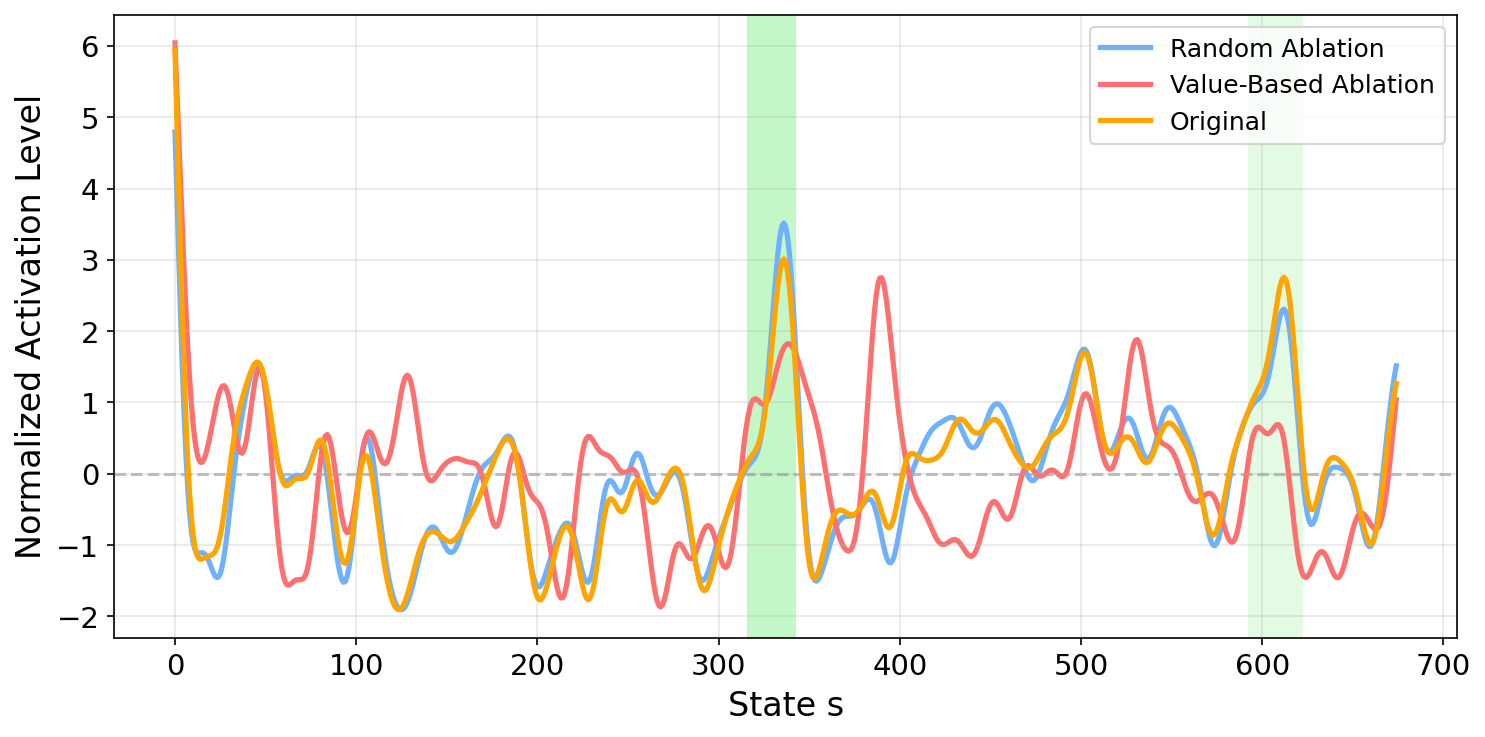}
    \end{minipage}
    \begin{minipage}{0.48\linewidth}
        \centering
        \includegraphics[width=\linewidth]{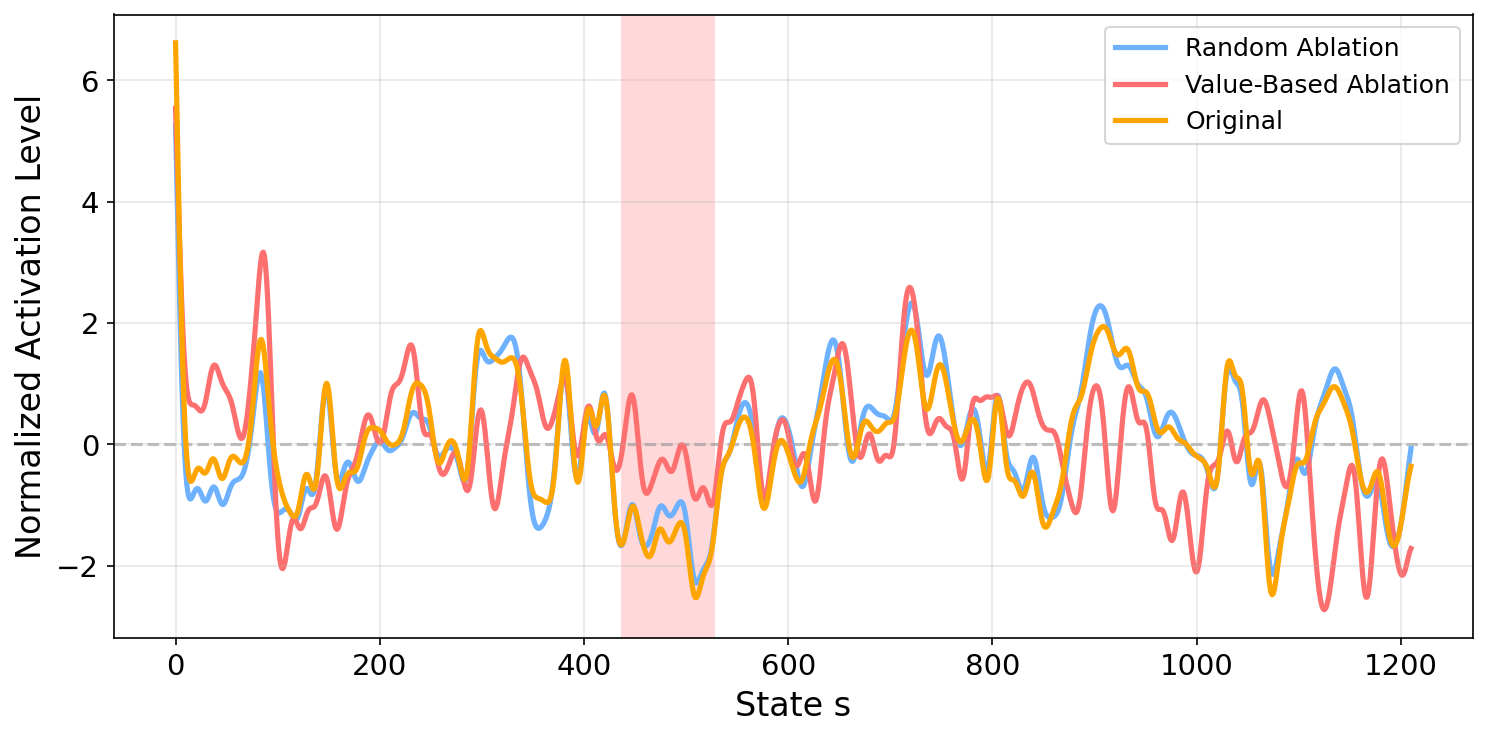}
    \end{minipage}
    \caption{
        The activation curves of a dopamine neuron across states under different ablation conditions. The yellow line represents the original trajectory, the blue line shows the result of zeroing 20\% random neurons, and the red line depicts the result of zeroing the top 20\% value neurons. While random ablation has a minimal impact on the overall trend, value neuron ablation significantly alters the trajectory of the curve. This indicates a close correlation between value and dopamine neurons.
    }
    \label{fig:abl_value_neurons} 
\end{figure}

We use the Qwen-2.5-14B-SimpleRL-Zoo model evaluated on the MATH500 dataset, and compare the effects of zeroing out the top 20\% of value neurons (identified by the highest $L_1$ norm in the value probe) versus zeroing out 20\% of randomly selected neurons. We then analyze the deviations in the normalized activation levels of the $1517$-th neuron in layer 5 relative to its original trajectory.

As shown in Figure~\ref{fig:abl_value_neurons}, while the blue line (random ablation) remains largely consistent with the original yellow curve aside from minor numerical fluctuations, the red line (value neuron ablation) exhibits fundamental differences. Specifically, the positions of the activation peaks and troughs shift significantly, causing the neuron to lose the characteristic properties of a dopamine neuron encoding prediction error. These results demonstrate that value and dopamine neurons are closely related, as disrupting a small subset of value neurons is sufficient to significantly impair the predictive performance of dopamine neurons.

\section{Value Neuron Analysis at the Final Response Position}
\label{app:last_token}

\begin{figure}[h]
\centering
\includegraphics[width=0.8\linewidth]{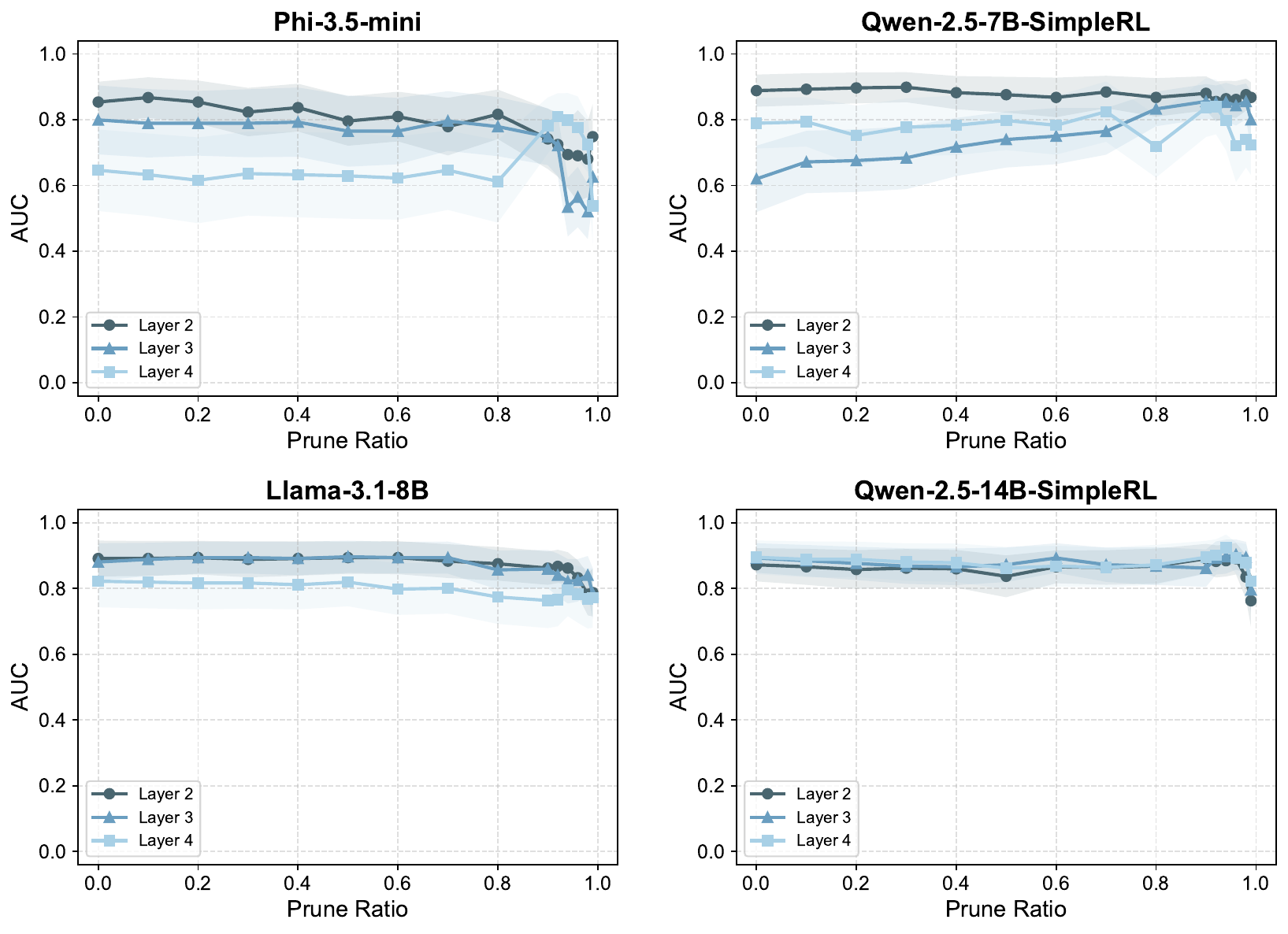}
\caption{AUC curves at the final response position for four models on the Minerva Math dataset. The AUC remains mostly stable above 0.8 across a wide range of pruning ratios. Shaded regions indicate uncertainty.}
\label{fig:last_token_auc}
\end{figure}

The experiments in the main paper evaluate value neurons using hidden states at the initial input position $s_0$, i.e., after the problem input but before the model begins generating its response. This setting tests whether the reward subsystem has already formed an assessment of the expected value prior to generation, but the signal at this position is relatively weak since the model has not yet begun reasoning.

A complementary setting is to evaluate value neurons at the last token of the model's completed response, where the reward-relevant signal can be expected to be stronger as the model has access to its full reasoning trajectory. We conduct this analysis using Phi-3.5-mini-instruct, Qwen-2.5-7B-SimpleRL-Zoo, Qwen-2.5-14B-SimpleRL-Zoo, and Llama-3.1-8B-Instruct on the Minerva Math~\citep{lewkowycz2022solving} dataset.

As shown in Figure~\ref{fig:last_token_auc}, the AUC remains mostly stable above 0.8, substantially higher than the pre-generation setting reported in the main paper. The consistency of this result across architectures and scales confirms that value neurons reliably encode reward information at both ends of the generation process, further strengthening the evidence for a sparse reward subsystem within LLM hidden states.

\section{Model Confidence Baseline Setup}
\label{app:baselines}

\paragraph{Verbal Confidence.}
In this straightforward baseline, we input the question into the LLM and use the prompt, \textit{'\textbackslash n\textbackslash n Rate your confidence in answering this question correctly from 0 to 1 (where 0 means no confidence and 1 means complete confidence). Only output a single number between 0 and 1. Do not try to solve the question.'} to elicit a confidence score. If the model fails to produce a valid numerical output, we resample until a score is obtained.

\paragraph{Token Confidence.}
In this baseline, we utilize the log probability ($\log p$) of the most likely token at the first generated position as a metric for the model's confidence. 

\paragraph{Latent Correctness Direction (LCD).}
We re-implement the method proposed by \citep{cencerrado2025answerneededpredictingllm} within our experimental setting. It is noteworthy that the baseline method utilizes the full set of neurons for prediction, so it cannot provide any guidance regarding the existence of the value neurons or the localization of neurons. In contrast, our approach utilizes significantly fewer hidden state dimensions than the baseline.

\section{Cross-Task Generalization of Trained Probes}
\label{app:cross_task_generalization}

The cross-dataset IoU analyses in Section~\ref{sec:transfer}
and Appendix~\ref{app:iou} show that value neurons identified
on different datasets overlap more than expected under
random selection. To complement these
analyses, we directly evaluate whether the trained probes transfer
across tasks. Specifically, we apply the value and dopamine probes
trained on MATH500 using Llama-3.1-8B-Instruct to GSM8K, evaluating
their performance on model confidence prediction and
process-reward-guided search, respectively.

\paragraph{Process-Reward-Guided Search.}
We use the transferred dopamine probe to score reasoning steps
and guide inference-time search on GSM8K. As shown in
Table~\ref{tab:cross_task_prm}, this approach achieves an accuracy
of 94.4\%, the highest mean accuracy among the evaluated methods.

\begin{table}[htbp]
    \centering
    \caption{Cross-task evaluation of process-reward-guided search.}
    \label{tab:cross_task_prm}
    \begin{tabular}{lc}
        \toprule
        Method & Accuracy (\%) \\
        \midrule
        Greedy & $90.7$ \\
        Random & $90.7$ \\
        Implicit PRM & $92.8$ \\
        Dopamine Neurons (MATH500-trained)
            & $\mathbf{94.4}$ \\
        \bottomrule
    \end{tabular}
\end{table}

\paragraph{Model Confidence Prediction.}
We also evaluate the transferred value probe for confidence
prediction on GSM8K. Table~\ref{tab:cross_task_confidence} shows
that it achieves an AUC of 0.692, the highest mean AUC among
the evaluated methods.

\begin{table}[htbp]
    \centering
    \caption{Cross-task evaluation of model confidence prediction. }
    \label{tab:cross_task_confidence}
    \begin{tabular}{lc}
        \toprule
        Method & AUC \\
        \midrule
        Token Confidence & $0.570$ \\
        Verbal Confidence & $0.640$ \\
        Question Length & $0.661$ \\
        LCD & $0.641$ \\
        Value Neurons (MATH500-trained)
            & $\mathbf{0.692}$ \\
        \bottomrule
    \end{tabular}
\end{table}

Together, these results provide evidence that the trained
value and dopamine probes retain predictive and practical utility
when transferred from MATH500 to GSM8K, supporting cross-task
generalization.

\section{Sensitivity to the TD-Error Threshold}
\label{app:threshold_ablation}

To examine the sensitivity to the TD-error threshold,
we vary $\tau$ in the filtering criterion $|\delta_t|>\tau$.
We conduct this ablation on Qwen-2.5-7B-SimpleRL-Zoo
with MATH500 across all layers, using three random seeds. Table~\ref{tab:threshold_ablation} reports the
mean Spearman correlation averaged across layers and seeds.
As shown in Table~\ref{tab:threshold_ablation}, $\tau=0.3$
achieves the highest mean Spearman correlation among the
evaluated thresholds, supporting the threshold used in
our experiments.
\begin{table}[h]
    \centering
    \caption{Mean Spearman correlation for different TD-error
    thresholds on Qwen-2.5-7B-SimpleRL-Zoo with MATH500,
    averaged across all layers. }
    \label{tab:threshold_ablation}
    \begin{tabular}{cc}
        \toprule
        \textbf{Threshold $\tau$}
        & \textbf{Mean $\rho$} \\
        \midrule
        0.1 & 0.152 \\
        0.2 & 0.267 \\
        0.3 & \textbf{0.281} \\
        0.4 & 0.193 \\
        \bottomrule
    \end{tabular}
\end{table}

\section{Is the Reward Subsystem Already Present in Base Models or Created by RL Fine-Tuning?}
\label{app:base_rl}

To examine whether value neurons are already present before
RL fine-tuning, we compare Qwen-2.5-7B with two RL-tuned
models derived from it: Qwen-2.5-7B-SimpleRL-Zoo and
Qwen2.5-7B-PPO-Zero. Both models are obtained through
RLVR using different training pipelines.

We train value probes on MATH500 for all three models
and compute the IoU between their identified value-neuron
sets at different pruning ratios $p$. The results are
averaged over layers 2--4 and compared against the random
baseline.
As shown in Table~\ref{tab:base_rl_iou}, the IoU between
the base model and both RL-tuned models consistently
exceeds the random baseline, even at high pruning ratios.
These results support the presence of value-related
representations in the base model before RL fine-tuning
and show substantial overlap in the identified coordinates
across two different RL pipelines.
\begin{table}[h]
    \centering
    \renewcommand{\arraystretch}{1.15}
    \caption{IoU between value-neuron sets identified in
    Qwen-2.5-7B and its two RL-tuned variants.
    Value probes are trained on MATH500, and results are
    averaged over layers 2--4. $p$ denotes the pruning ratio.}
    \label{tab:base_rl_iou}
    \begin{tabular*}{\textwidth}{
        c @{\extracolsep{\fill}} cccc}
        \toprule
        \makecell{$p$}
        & \makecell{Random\\Baseline}
        & \makecell{Base vs\\SimpleRL-Zoo}
        & \makecell{Base vs\\PPO-Zero}
        & \makecell{SimpleRL-Zoo\\vs PPO-Zero} \\
        \midrule
        0.50 & 0.333 & 0.723 & 0.705 & 0.676 \\
        0.80 & 0.111 & 0.513 & 0.525 & 0.454 \\
        0.90 & 0.053 & 0.429 & 0.408 & 0.362 \\
        0.96 & 0.020 & 0.326 & 0.316 & 0.266 \\
        0.98 & 0.010 & 0.290 & 0.310 & 0.235 \\
        0.99 & 0.005 & 0.283 & 0.290 & 0.236 \\
        \bottomrule
    \end{tabular*}
\end{table}

\section{Additional Discussion}
\label{app:discussion}

\paragraph{Internal Estimates of Success and Failure.}
Our results suggest that sparse hidden-state coordinates
carry information about the expected success of the current
reasoning state. Value-neuron probes outperform verbalized
confidence on average, suggesting that internal representations
contain success-related information that is not fully captured
by verbalized confidence. This motivates investigating whether
changes in value estimates can anticipate errors as reasoning
unfolds.

\paragraph{Training Interventions and Steering.}
Value and dopamine probes offer lightweight value and
process-reward signals that may guide post-training and
inference-time search. Because these probes estimate
reward-related quantities under the current policy, they
may require refreshing when the policy changes substantially.
A natural next step is to investigate targeted steering of
the identified coordinates and measure its effects on both
probe predictions and actual task performance.

\paragraph{Relationship to the Implicit Q-Function View.}
The policy--value relationship in
Equation~\ref{eq:maxent_log_policy} motivates searching for
value-related information within policy representations.
Our findings complement this view by showing that state
value and step-level TD-error information can be predicted
from sparse hidden-state coordinates. 

\paragraph{Potential Connections to Evaluation-Aware Behavior.}
Internal estimates of expected reward may provide a useful
perspective for studying evaluation awareness and reward
seeking~\citep{evalaware}. If models use such estimates when adapting their
behavior to evaluation contexts, the identified coordinates
could offer signals for investigating these adaptations.
Another possible application is lightweight monitoring:
discrepancies between internal value estimates and overt
behavior might help identify evaluation-aware behavior.
These connections remain hypotheses requiring direct
evaluation.

\paragraph{Scope of the Neuroscience Analogy.}
The terms ``value neurons'' and ``dopamine neurons'' describe
analogies in computational role: predicting state value and
reward prediction errors, respectively. Here, neurons are
residual-stream coordinates selected through supervised
probing, rather than biologically characterized cell types.
The analogy therefore guides interpretation and future
experiments without establishing a biological correspondence.

\section{Potential Limitations and Broader Impacts}
\label{app:lim}

\textbf{Limitations.} We acknowledge the following limitations of the paper:

(1) While our benchmarks cover a diverse range of tasks including mathematics, reasoning, coding, and general instruction following, our method requires reward signals from the environment and has not been tested on tasks where such signals are difficult to obtain, such as open-ended generation.

(2) Due to resource and computational constraints, we have not investigated whether similar phenomena exist in models larger than 32B.

(3) Our intervention experiments offer preliminary insights, and potential confounds, such as probe stability across rollouts and large activation magnitudes, remain for future work. We thank Nell Watson for her interest in our work and for her independent pre-registered replication, which provided valuable insights and improved our understanding of these results.

\textbf{Broader Impacts.} 
Our work identifies sparse neurons that encode reward-related information in LLMs, which has implications for both interpretability and control. On the positive side, understanding where reward information resides can improve transparency in LLM reasoning. However, this capability also raises concerns. Neuron-level interventions on reward subsystems could potentially be used to manipulate model behavior in unintended ways. We encourage future work to investigate the robustness of these neurons under adversarial settings and to develop safeguards before deploying neuron-level control mechanisms in practice.

%%%%%%%%%%%%%%%%%%%%%%%%%%%%%%%%%%%%%%%%%%%%%%%%%%%%%%%%%%%%

\newpage
\section*{NeurIPS Paper Checklist}

\begin{enumerate}

\item {\bf Claims}
    \item[] Question: Do the main claims made in the abstract and introduction accurately reflect the paper's contributions and scope?
    \item[] Answer: \answerYes{} % Replace by \answerYes{}, \answerNo{}, or \answerNA{}.
    \item[] Justification: We clearly state our claims in the abstract and the introduction section. They accurately reflect the methods and the experiments.

\item {\bf Limitations}
    \item[] Question: Does the paper discuss the limitations of the work performed by the authors?
    \item[] Answer: \answerYes{} % Replace by \answerYes{}, \answerNo{}, or \answerNA{}.
    \item[] Justification: We mentioned potential limitations in Appendix \ref{app:lim}.

\item {\bf Theory assumptions and proofs}
    \item[] Question: For each theoretical result, does the paper provide the full set of assumptions and a complete (and correct) proof?
    \item[] Answer: \answerNA{} % Replace by \answerYes{}, \answerNo{}, or \answerNA{}.
    \item[] Justification: The paper does not include formal theoretical results. 

    \item {\bf Experimental result reproducibility}
    \item[] Question: Does the paper fully disclose all the information needed to reproduce the main experimental results of the paper to the extent that it affects the main claims and/or conclusions of the paper (regardless of whether the code and data are provided or not)?
    \item[] Answer: \answerYes{} % Replace by \answerYes{}, \answerNo{}, or \answerNA{}.
    \item[] Justification:  We clearly describe our methods in Section \ref{sec:value} and Section \ref{sec:dopamine}. 

\item {\bf Open access to data and code}
    \item[] Question: Does the paper provide open access to the data and code, with sufficient instructions to faithfully reproduce the main experimental results, as described in supplemental material?
    \item[] Answer: \answerYes{} % Replace by \answerYes{}, \answerNo{}, or \answerNA{}.
    \item[] Justification:  We provide open code access.

\item {\bf Experimental setting/details}
    \item[] Question: Does the paper specify all the training and test details (e.g., data splits, hyperparameters, how they were chosen, type of optimizer) necessary to understand the results?
    \item[] Answer: \answerYes{} % Replace by \answerYes{}, \answerNo{}, or \answerNA{}.
    \item[] Justification: The experimental settings are provided in Appendix \ref{app:hyper} and Appendix \ref{app:baselines}.

\item {\bf Experiment statistical significance}
    \item[] Question: Does the paper report error bars suitably and correctly defined or other appropriate information about the statistical significance of the experiments?
    \item[] Answer: \answerYes{} % Replace by \answerYes{}, \answerNo{}, or \answerNA{}.
    \item[] Justification: We compute all curves via bootstrap resampling on the validation set. 

\item {\bf Experiments compute resources}
    \item[] Question: For each experiment, does the paper provide sufficient information on the computer resources (type of compute workers, memory, time of execution) needed to reproduce the experiments?
    \item[] Answer: \answerYes{} % Replace by \answerYes{}, \answerNo{}, or \answerNA{}.
    \item[] Justification: The experiments compute resources are provided in Appendix \ref{app:hyper}.
    
\item {\bf Code of ethics}
    \item[] Question: Does the research conducted in the paper conform, in every respect, with the NeurIPS Code of Ethics \url{https://neurips.cc/public/EthicsGuidelines}?
    \item[] Answer: \answerYes{} % Replace by \answerYes{}, \answerNo{}, or \answerNA{}.
    \item[] Justification: The paper conforms with the NeurIPS code of ethics.

\item {\bf Broader impacts}
    \item[] Question: Does the paper discuss both potential positive societal impacts and negative societal impacts of the work performed?
    \item[] Answer: \answerYes{} % Replace by \answerYes{}, \answerNo{}, or \answerNA{}.
    \item[] Justification: We discuss the broader impacts in Appendix \ref{app:lim}.
    
\item {\bf Safeguards}
    \item[] Question: Does the paper describe safeguards that have been put in place for responsible release of data or models that have a high risk for misuse (e.g., pre-trained language models, image generators, or scraped datasets)?
    \item[] Answer: \answerNA{} % Replace by \answerYes{}, \answerNo{}, or \answerNA{}.
    \item[] Justification: The paper poses no such risks.

\item {\bf Licenses for existing assets}
    \item[] Question: Are the creators or original owners of assets (e.g., code, data, models), used in the paper, properly credited and are the license and terms of use explicitly mentioned and properly respected?
    \item[] Answer: \answerYes{} % Replace by \answerYes{}, \answerNo{}, or \answerNA{}.
    \item[] Justification: They are properly credited, mentioned and respected.

\item {\bf New assets}
    \item[] Question: Are new assets introduced in the paper well documented and is the documentation provided alongside the assets?
    \item[] Answer:  \answerYes{} % Replace by \answerYes{}, \answerNo{}, or \answerNA{}.
    \item[] Justification: Yes, code is the only new asset. It is  well documented and the documentation is  provided alongside the code.

\item {\bf Crowdsourcing and research with human subjects}
    \item[] Question: For crowdsourcing experiments and research with human subjects, does the paper include the full text of instructions given to participants and screenshots, if applicable, as well as details about compensation (if any)? 
    \item[] Answer: \answerNA{} % Replace by \answerYes{}, \answerNo{}, or \answerNA{}.
    \item[] Justification: The paper does not involve crowdsourcing nor research with human subjects.

\item {\bf Institutional review board (IRB) approvals or equivalent for research with human subjects}
    \item[] Question: Does the paper describe potential risks incurred by study participants, whether such risks were disclosed to the subjects, and whether Institutional Review Board (IRB) approvals (or an equivalent approval/review based on the requirements of your country or institution) were obtained?
    \item[] Answer: \answerNA{} % Replace by \answerYes{}, \answerNo{}, or \answerNA{}.
    \item[] Justification: The paper does not involve crowdsourcing nor research with human subjects.

\item {\bf Declaration of LLM usage}
    \item[] Question: Does the paper describe the usage of LLMs if it is an important, original, or non-standard component of the core methods in this research? Note that if the LLM is used only for writing, editing, or formatting purposes and does \emph{not} impact the core methodology, scientific rigor, or originality of the research, declaration is not required.
    %this research? 
    \item[] Answer: \answerNA{} % Replace by \answerYes{}, \answerNo{}, or \answerNA{}.
    \item[] Justification: The core method development in this research does not involve LLMs as any important, original, or non-standard components.

\end{enumerate}

\end{document}